arXiv:1701.05672v1 [stat.ML] 20 Jan 2017

# Stability Enhanced Large-Margin Classifier Selection

Will Wei Sun[*], Guang Cheng[†], Yufeng Liu[‡]

**Abstract**

Stability is an important aspect of a classification procedure because unstable predictions can potentially reduce users' trust in a classification system and also harm the reproducibility of scientific conclusions. The major goal of our work is to introduce a novel concept of classification instability, i.e., decision boundary instability (DBI), and incorporate it with the generalization error (GE) as a standard for selecting the *most accurate and stable* classifier. Specifically, we implement a two-stage algorithm: (i) initially select a subset of classifiers whose estimated GEs are not significantly different from the minimal estimated GE among all the candidate classifiers; (ii) the optimal classifier is chosen as the one achieving the minimal DBI among the subset selected in stage (i). This general selection principle applies to both linear and nonlinear classifiers. Large-margin classifiers are used as a prototypical example to illustrate the above idea. Our selection method is shown to be consistent in the sense that the optimal classifier simultaneously achieves the minimal GE and the minimal DBI. Various simulations and real examples further demonstrate the advantage of our method over several alternative approaches.

## 1 Introduction

Classification aims to identify the class label of a new subject using a classifier constructed from training data whose class memberships are given. It has been widely used in diverse fields, e.g., medical diagnosis, fraud detection, and natural language processing. Numerous classification methods have been successfully developed with classical approaches such as Fisher's linear discriminant analysis (LDA), quadratic discriminant analysis (QDA), and logistic regression (see Hastie et al., 2001 for a comprehensive review), and modern approaches such as the support vector machine (SVM) (Cortes and Vapnik, 1995) and boosting (Freund and Schapire, 1997). In a recent paper, Liu et al. (2011) proposed a platform, large-margin unified machines (LUM), for unifying various large margin classifiers ranging from soft to hard.

[*]Assistant Professor, Department of Management Science, University of Miami, FL 33156, Email: wsun@bus.miami.edu. This work was carried out during Will's PhD period at Purdue University.

[†]Corresponding Author. Professor, Department of Statistics, Purdue University, West Lafayette, IN 47906, Email: chengg@purdue.edu. Partially supported by NSF CAREER Award DMS-1151692, DMS-1418042 and Office of Naval Research (ONR N00014-15-1-2331). I acknowledge SAMSI and Princeton University for their hospitality; part of the work was done during my visits.

[‡]Professor, Department of Statistics and Operations Research, Department of Genetics, Department of Biostatistics, Carolina Center for Genome Sciences, Lineberger Comprehensive Cancer Center, University of North Carolina Chapel Hill, NC 27599, Email: yfliu@email.unc.edu. Partially supported by NSF DMS-1407241, and NIH/NCI P01 CA-142538.

In the literature, much of the research has focused on improving the predictive accuracy of classifiers and hence generalization error (GE) is often the primary criterion for selecting the optimal one from the rich pool of existing classifiers; see Vapnik (1998) and Steinwart (2007). Recently, researchers have started to explore alternative measures to evaluate the performance of classifiers. For instance, besides prediction accuracy, computational complexity and training time of classifiers are considered in Lim et al. (2000). Moreover, Wu and Liu (2007) proposed the robust truncated hinge loss SVM to improve the robustness of the standard SVM. Qiao and Liu (2009) and Wang (2013) investigated several measures of cost-sensitive weighted generalization errors for highly unbalanced classification tasks since, in this case, GE itself is not sufficiently informative. In this paper, we focus on the stability of a classification procedure. In general, stability has received well-deserved attention in statistics and machine learning. For example, Wang (2010) employed clustering instability as a criterion to select the number of clusters; Adomavicius and Zhang (2010) introduced stability as a new performance measure for recommender systems; Meinshausen and Bühlmann (2010) and Shah and Samworth (2013) used stability for variable selection; Sun et al. (2013) applied variable selection stability for model selection, and Lim and Yu (2016) incorporated estimation stability into the tuning parameter selection of regularized regression models. While successes of stability have been reported in the aforementioned works, little work has been developed for classification stability itself, expect for a recent one on nearest neighbor classifiers (Sun et al., 2016). Consequently, there is a great need for a systematic study of stability in a general classification context.

In this article, we introduce a notion of decision boundary instability (DBI) to assess the stability (Breiman, 1996) of a classification procedure arising from the randomness of training samples. Stability is an important aspect of a classification procedure. First, providing a stable prediction plays a crucial role on users' trust of the classification system. In the psychology literature, it has been shown that advice-giving agents with lager variability in past opinions are considered less informative and less helpful than those with a more consistent pattern of opinions (Gershoff et al., 2003; Van Swol and Sniezek, 2005). Therefore, a classification procedure may be distrusted by users if it generates highly unstable predictions simply due to the randomness of training samples. Second, scientific conclusions should be reproducible with respect to small perturbation of data. Reproducible research has recently received much attention in statistics (Yu, 2013), biostatistics (Kraft et al., 2009; Peng, 2009), computational science (Donoho et al., 2009) and other scientific communities (Ioannidis, 2005). A classification procedure with more stable prediction performance is preferred when researchers aim to reproduce the reported results from randomly generated samples. Consequently, aside from high prediction accuracy, high stability is another crucial factor to consider in the classifier selection.

In this paper, we attempt to select the *most accurate and stable* classifier by incorporating DBI into our selection process. Specifically, we suggest a two-stage selection procedure: (i) eliminate the classifiers whose GEs are significantly larger than the minimal one among all the candidate classifiers; (ii) select the optimal classifier as that with the most stable decision boundary, i.e., the minimal DBI, among the remaining classifiers.

In the first stage, we show that the cross-validation estimator for the difference of GEs induced from two large-margin classifiers asymptotically follows Gaussian distribution, which enables us to construct a confidence interval for the GE difference. If this confidence interval contains 0, these two classifiers are considered indistinguishable in terms of GE. By applying the above approach, we can obtain a collection of potentially good classifiers whose GEs are close enough to the minimal value. The uncertainty quantification of the cross-validation estimator is crucially important considering that only limited samples are available in practice. In fact, experiments indicate that for certain problems many classifiers do not significantly differ in their estimated GEs, and the corresponding absolute differences are mainly due to random noise. In the second stage, we propose to check whether the collection of potentially good classifiers also perform well in terms of their stability. We observe that the decision boundary generated by the classifier with the minimal GE estimator sometimes has unstable behavior given a small perturbation of the training samples. This observation motivates us to propose a further selection criterion: DBI. This new measure can precisely reflect the visual variability in the decision boundaries due to the perturbed training samples.

Our two-stage selection algorithm is shown to be consistent in the sense that the selected optimal classifier simultaneously achieves the minimal GE and the minimal DBI. The proof is nontrivial because of the stochastic nature of the two-stage algorithm. Note that our method is distinguished from the bias-variance analysis in classification since the latter focuses on the decomposition of GE, e.g., Valentini and Dietterich (2004). Our DBI is also conceptually different from the stability-oriented measure introduced in Bousquet and Elisseeff (2002), which was defined as the maximal difference of the decision functions trained from the original datasets and the leave-one-out datasets. In addition, their variability measure suffers from the transformation variant issue since a scale transformation of the decision function coefficients will greatly affect their variability measure. Our DBI overcomes this problem via a rescaling scheme since DBI can be viewed as a weighted version of the asymptotic variance of the decision function. More discussions on the connection with other variability measures are given in Section 3.3. In the end, extensive experiments illustrate the advantage of our selection algorithm compared with the alternative approaches in terms of both classification accuracy and stability.

For simplicity, we focus on linear classifiers in this paper. The nonlinear extension becomes conceptually feasible by mapping the nonlinear feature space into a higher dimensional linear space; see Appendix for further discussion. The rest of the article is organized as follows. Section 2 reviews large-margin classifiers, which are used as prototypical examples to illustrate our method. Section 3 describes the main results of our classifier selection procedure. Section 4 establishes the selection consistency of the proposed selection procedure. The simulated and real examples are in Section 5, followed by a brief discussion in Section 6. The Appendix and Supplementary Materials are devoted to the technical details and a notation table.

## 2 Large-Margin Classifiers

This section briefly reviews large-margin classifiers, which serve as prototypical examples to illustrate our two-stage classifier selection technique. It is worth noting that the proposed method is broadly applicable to general classifiers.

Let $(\boldsymbol{X}, Y) \in \mathbb{R}^d \times \{1, -1\}$ be random variables from an underlying distribution $\mathcal{P}(\boldsymbol{X}, Y)$. Denote the conditional probability of class $Y = 1$ given $\boldsymbol{X} = \boldsymbol{x}$ as $p(\boldsymbol{x}) = P(Y = 1|\boldsymbol{X} = \boldsymbol{x})$, where $p(\boldsymbol{x}) \in (0, 1)$ to exclude the degenerate case. Let the input variable be $\boldsymbol{x} = (x_1, \ldots, x_d)^T$, $\tilde{\boldsymbol{x}} = (1, x_1, \ldots, x_d)^T$, with coefficient $\boldsymbol{w} = (w_1, \ldots, w_d)^T$ and parameter $\boldsymbol{\theta} = (b, \boldsymbol{w}^T)^T$. The lin-ear decision function is defined as $f(\boldsymbol{x};\boldsymbol{\theta}) = b + \boldsymbol{x}^T\boldsymbol{w} = \tilde{\boldsymbol{x}}^T\boldsymbol{\theta}$, and the decision boundary is $S(\boldsymbol{x};\boldsymbol{\theta}) = \{\boldsymbol{x} : f(\boldsymbol{x};\boldsymbol{\theta}) = 0\}$. The performance of the classifier $\text{sign}\{f(\boldsymbol{x};\boldsymbol{\theta})\}$ is measured by the classification risk $E[\mathbb{1}\{Y \neq \text{sign}\{f(\boldsymbol{X};\boldsymbol{\theta})\}\}]$, where the expectation is with respect to $\mathcal{P}(\boldsymbol{X}, Y)$. Since the direct minimization of the above risk is NP hard (Zhang, 2004), various convex surrogate loss functions $L(\cdot)$ have been proposed to deal with this computational issue. Denote the surrogate risk as $\mathcal{R}_L(\boldsymbol{\theta}) = E[L(Yf(\boldsymbol{X};\boldsymbol{\theta}))]$, and assume that the minimizer of $\mathcal{R}_L(\boldsymbol{\theta})$ is obtained at $\boldsymbol{\theta}_{0L} = (b_{0L}, \boldsymbol{w}_{0L}^T)^T$. Here $\boldsymbol{\theta}_{0L}$ depends on the loss function $L$.

Given the training sample $\mathcal{D}_n = \{(\boldsymbol{x}_i, y_i); i = 1, \ldots, n\}$ drawn from $\mathcal{P}(\boldsymbol{X}, Y)$, a large-margin classifier minimizes the empirical risk $O_{nL}(\boldsymbol{\theta})$ defined as

$$O_{nL}(\boldsymbol{\theta}) = \frac{1}{n}\sum_{i=1}^{n} L\Big(y_i(\boldsymbol{w}^T\boldsymbol{x}_i + b)\Big) + \frac{\lambda_n}{2}\boldsymbol{w}^T\boldsymbol{w}, \tag{1}$$

where $\lambda_n$ is some positive tuning parameter. The estimator minimizing $O_{nL}(\boldsymbol{\theta})$ is denoted as $\widehat{\boldsymbol{\theta}}_L = (\widehat{b}_L, \widehat{\boldsymbol{w}}_L^T)^T$. Common large-margin classifiers include the squared loss $L(u) = (1-u)^2$, the exponential loss $L(u) = e^{-u}$, the logistic loss $L(u) = \log(1+e^{-u})$, and the hinge loss $L(u) = (1-u)_+$. Unfortunately, there seems to be no general guideline for selecting these loss functions in practice except the cross validation error. Ideally if we had access to an arbitrarily large test set, we would just choose the classifier for which the test error is the smallest. However, in reality where only limited samples are available, the commonly used cross validation error may not be able to accurately approximate the testing error. The main goal of this paper is to establish a practically useful selection criterion by incorporating DBI with the cross validation error.

# 3 Classifier Selection Algorithm

In this section, we propose a two-stage classifier selection algorithm: (i) we select candidate classifiers whose estimated GEs are relatively small; (ii) the optimal classifier is that with the minimal DBI among those selected in Stage (i).

## 3.1 Stage 1: Initial Screening via GE

In this subsection, we show that the difference of the cross-validation errors obtained from two large-margin classifiers asymptotically follows Gaussian distribution, which enables us to construct a confidence interval for their GE difference. We further propose a perturbation-based resampling approach to construct this confidence interval.

Given a new input $(\boldsymbol{X}_0, Y_0)$ from $\mathcal{P}(\boldsymbol{X}, Y)$, we define the GE induced by the loss function $L$ as

$$D_{0L} = \frac{1}{2}E|Y_0 - \text{sign}\{f(\boldsymbol{X}_0;\widehat{\boldsymbol{\theta}}_L)\}|, \tag{2}$$

where $\widehat{\boldsymbol{\theta}}_L$ is based on the training sample $\mathcal{D}_n$, and the expectation is with respect to both $\mathcal{D}_n$ and $(\boldsymbol{X}_0, Y_0)$. Note that GE in (2) is equivalent to the mis-classification risk $E[\mathbb{1}\{Y_0 \neq \text{sign}\{f(\boldsymbol{X}_0;\widehat{\boldsymbol{\theta}}_L)\}\}]$. In practice, the GE, which depends on the underlying distribution $\mathcal{P}(\boldsymbol{X}, Y)$, needs to be estimated using $\mathcal{D}_n$. One possible estimate is the empirical generalization error defined as $\widehat{D}_L \equiv \widehat{D}(\widehat{\boldsymbol{\theta}}_L)$, where $\widehat{D}(\boldsymbol{\theta}) = (2n)^{-1}\sum_{i=1}^{n}|y_i - \text{sign}\{f(\boldsymbol{x}_i;\boldsymbol{\theta})\}|$. However, the above estimate suffers from the problem

of overfitting (Wang and Shen, 2006). Hence, one can use the K-fold cross-validation procedure to estimate the GE; this can significantly reduce the bias (Jiang et al., 2008). Specifically, we randomly split $\mathcal{D}_n$ into $K$ disjoint subgroups and denote the $k$th subgroup as $I_k$. For $k = 1, \dots, K$, we obtain the estimator $\widehat{\boldsymbol{\theta}}_{L(-k)}$ from all the data except those in $I_k$, and calculate the empirical average $\widehat{D}(\widehat{\boldsymbol{\theta}}_{L(-k)})$ based only on $I_k$, i.e., $\widehat{D}(\widehat{\boldsymbol{\theta}}_{L(-k)}) = (2|I_k|)^{-1} \sum_{i \in I_k} |y_i - \text{sign}\{f(\boldsymbol{x}_i; \widehat{\boldsymbol{\theta}}_{L(-k)})\}|$ with $|I_k|$ being the cardinality of $I_k$. The K-fold cross-validation (K-CV) error is thus computed as

$$\widehat{\mathcal{D}}_L = K^{-1} \sum_{k=1}^{K} \widehat{D}(\widehat{\boldsymbol{\theta}}_{L(-k)}). \tag{3}$$

We set $K = 5$ for our numerical experiments.

In order to establish the asymptotic normality of the K-CV error $\widehat{\mathcal{D}}_L$ for a general loss $L(\cdot)$, we require the following regularity conditions on the population distribution and the loss function.

(L1) The probability distribution function of $\boldsymbol{X}$ and the conditional probability $p(\boldsymbol{x})$ are both continuously differentiable.

(L2) The parameter $\boldsymbol{\theta}_{0L}$ is bounded and unique.

(L3) The map $\boldsymbol{\theta} \mapsto L(yf(\boldsymbol{x}; \boldsymbol{\theta}))$ is convex.

(L4) The map $\boldsymbol{\theta} \mapsto L(yf(\boldsymbol{x}; \boldsymbol{\theta}))$ is differentiable at $\boldsymbol{\theta} = \boldsymbol{\theta}_{0L}$ a.s.. Furthermore, $G(\boldsymbol{\theta}_{0L})$ is elementwisely bounded, where

$$G(\boldsymbol{\theta}_{0L}) = E\Big[\nabla_{\boldsymbol{\theta}} L(Yf(\boldsymbol{X}; \boldsymbol{\theta})) \nabla_{\boldsymbol{\theta}} L(Yf(\boldsymbol{X}; \boldsymbol{\theta}))^T\Big]\Big|_{\boldsymbol{\theta} = \boldsymbol{\theta}_{0L}}.$$

(L5) The surrogate risk $\mathcal{R}_L(\boldsymbol{\theta})$ is bounded and twice differentiable at $\boldsymbol{\theta} = \boldsymbol{\theta}_{0L}$ with the positive definite Hessian matrix $H(\boldsymbol{\theta}_{0L}) = \nabla^2_{\boldsymbol{\theta}} \mathcal{R}_L(\boldsymbol{\theta})|_{\boldsymbol{\theta} = \boldsymbol{\theta}_{0L}}$.

Assumption (L1) ensures that the GE is continuously differentiable with respect to $\boldsymbol{\theta}$ so that the uniform law of large numbers can be applied. Assumption (L3) ensures that the uniform convergence theorem for convex functions (Pollard, 1991) can be applied, and it is satisfied by all the large-margin loss functions considered in this paper. Assumptions (L4) and (L5) are required to obtain the local quadratic approximation to the surrogate risk function around $\boldsymbol{\theta}_{0L}$. Assumptions (L2)–(L5) were previously used by Rocha et al. (2009) to prove the asymptotic normality of $\widehat{\boldsymbol{\theta}}_L$.

Theorem 1 below establishes the asymptotic normality of the K-CV error $\widehat{\mathcal{D}}_L$ for any large-margin classifier, which generalizes the result for the SVM in Jiang et al. (2008).

**Theorem 1** *Suppose Assumptions (L1)–(L5) hold and $\lambda_n = o(n^{-1/2})$. Then for any fixed $K$,*

$$\mathcal{W}_L = \sqrt{n}\Big(\widehat{\mathcal{D}}_L - D_{0L}\Big) \xrightarrow{d} N\Big(0, E(\psi_1^2)\Big) \quad \text{as } n \to \infty, \tag{4}$$

*where $\psi_1 = \frac{1}{2}|Y_1 - \text{sign}\{f(\boldsymbol{X}_1; \boldsymbol{\theta}_{0L})\}| - D_{0L} - \dot{d}(\boldsymbol{\theta}_{0L})^T H(\boldsymbol{\theta}_{0L})^{-1} M_1(\boldsymbol{\theta}_{0L})$ with $\dot{d}(\boldsymbol{\theta}) = \nabla_{\boldsymbol{\theta}} E(\widehat{D}(\boldsymbol{\theta}))$, and $M_1(\boldsymbol{\theta}) = \nabla_{\boldsymbol{\theta}} L(Y_1 f(\boldsymbol{X}_1; \boldsymbol{\theta}))$.*

The proof of Theorem 1 is included in Section S.1 of the online supplement. An immediate application of Theorem 1 is to compare two competing loss functions $L_1$ and $L_2$. Define their GE difference $\Delta_{12}$ and its consistent estimate $\widehat{\Delta}_{12}$ to be $D_{02} - D_{01}$ and $\widehat{\mathcal{D}}_2 - \widehat{\mathcal{D}}_1$, respectively. To test whether the GEs induced by $L_1$ and $L_2$ are significantly different, we need to establish an approximate confidence interval for $\Delta_{12}$ based on the distribution of $\mathcal{W}_{\Delta_{12}} \equiv \mathcal{W}_2 - \mathcal{W}_1 = n^{1/2}(\widehat{\Delta}_{12} - \Delta_{12})$. In practice, we apply the perturbation-based resampling procedure (Park and Wei, 2003) to approximate the distribution of $\mathcal{W}_{\Delta_{12}}$. This procedure was also employed by Jiang et al. (2008) to construct the confidence interval of SVM's GE. Specifically, let $\{G_i\}_{i=1}^n$ be i.i.d. random variables drawn from the exponential distribution with unit mean and unit variance. Denote

$$\widehat{\boldsymbol{\theta}}_j^* = \arg\min_{b,\boldsymbol{w}} \left\{ \frac{1}{n} \sum_{i=1}^n G_i L_j\Big(y_i(\boldsymbol{w}^T \boldsymbol{x}_i + b)\Big) + \frac{\lambda_n}{2} \boldsymbol{w}^T \boldsymbol{w} \right\}. \tag{5}$$

Conditionally on $\mathcal{D}_n$, the randomness of $\widehat{\boldsymbol{\theta}}_j^*$ merely comes from that of $G_1, \ldots, G_n$. Denote $W_{\Delta_{12}}^* = W_2^* - W_1^*$ with

$$W_j^* = n^{-1/2} \sum_{i=1}^n \left\{ \frac{1}{2} \Big| y_i - \mathrm{sign}\{f(\boldsymbol{x}_i, \widehat{\boldsymbol{\theta}}_j^*)\} \Big| - \widehat{D}_j \right\} G_i. \tag{6}$$

By repeatedly generating a set of random variables $\{G_i, i = 1, \ldots, n\}$, we can obtain a large number of realizations of $W_{\Delta_{12}}^*$ to approximate the distribution of $\mathcal{W}_{\Delta_{12}}$. In Theorem 2 below, we prove that this approximation is valid. Its proof is included in Section S.2 of the online supplement.

**Theorem 2** *Suppose that the assumptions in Theorem 1 hold. Then as $n \to \infty$,*

$$\mathcal{W}_{\Delta_{12}} \xrightarrow{d} N\Big(0, Var(\psi_{12} - \psi_{11})\Big),$$

*where $\psi_{11}$ and $\psi_{12}$ are defined in Section S.2 of the online supplement, and*

$$W_{\Delta_{12}}^* \overset{d}{\Longrightarrow} N\Big(0, Var(\psi_{12} - \psi_{11})\Big) \quad \textit{conditional on } \mathcal{D}_n,$$

*where "$\Longrightarrow$" means conditional weak convergence in the sense of Hoffmann-Jorgensen (1984).*

*Algorithm 1* below summarizes the resampling procedure for establishing the confidence interval of the GE difference $\Delta_{12}$.

*Algorithm 1 (Generalization Error Comparison Algorithm)*

Input: Training sample $\mathcal{D}_n$ and two candidate loss functions $L_1$ and $L_2$.

- Step 1. Calculate K-CV errors $\widehat{\mathcal{D}}_1$ and $\widehat{\mathcal{D}}_2$ induced from $L_1$ and $L_2$, respectively.
- Step 2. For $r = 1, \ldots, N$, repeat the following steps:
  - (a) Generate i.i.d. samples $\{G_i^{(r)}\}_{i=1}^n$ from Exp(1);
  - (b) Find $\widehat{\boldsymbol{\theta}}_j^{*(r)}$ via (5) and $W_j^{*(r)}$ via (6), and calculate $W_{\Delta_{12}}^{*(r)} = W_2^{*(r)} - W_1^{*(r)}$.
- Step 3. Construct the $100(1-\alpha)\%$ confidence interval for $\Delta_{12}$ as

  $$\Big[\widehat{\Delta}_{12} - n^{-1/2}\phi_{1,2;\alpha/2}, \widehat{\Delta}_{12} - n^{-1/2}\phi_{1,2;1-\alpha/2}\Big],$$

  where $\widehat{\Delta}_{12} = \widehat{\mathcal{D}}_2 - \widehat{\mathcal{D}}_1$ and $\phi_{1,2;\alpha}$ is the $\alpha$th upper percentile of $\{W_{\Delta_{12}}^{*(1)}, \ldots, W_{\Delta_{12}}^{*(N)}\}$.

In our experiments, we repeated the resampling procedure 100 times, i.e., $N = 100$ in Step 2, and fix $\alpha = 0.1$. The effect of the choice of $\alpha$ will be discussed at the end of Section 3.4. The GEs of two classifiers induced from $L_1$ and $L_2$ are significantly different if the confidence interval established in Step 3 does not contain 0. Hence, we can apply *Algorithm 1* to eliminate the classifiers whose GEs are significantly different from the minimal GE of a set of candidate classifiers.

It is worth noting that employing statistical testing for classifier comparison has been successfully applied in practice (Dietterich, 1998; Demsar, 2006). In particular, Demsar (2006) reviewed several statistical tests in comparing two classifiers on multiple data sets and recommended the Wilcoxon sign rank test, which examined whether two classifiers are significantly different by calculating the relative rank of their corresponding performance scores on multiple data sets. Their result relies on an ideal assumption that there is no sampling variability of the measured performance score in each individual data set. Compared to the Wilcoxon sign rank test, our perturbed cross validation estimator has the advantages that it is theoretically justified and it does not rely on the ideal assumption of each performance score.

The remaining classifiers from *Algorithm 1* are potentially good. As will be seen in the next section, the decision boundaries of potentially good classifiers may change dramatically following a small perturbation of the training sample. This indicates that the prediction stability of the classifiers can be different although their GEs are fairly close. Motivated by this observation, in the next section we introduce the DBI to capture the prediction instability and embed it into our classifier selection algorithm.

### 3.2 Stage 2: Final Selection via DBI

In this section, we define the DBI and then provide an efficient way to estimate it in practice.

**Toy Example:** To motivate the DBI, we start with a simulated example using two classifiers: the squared loss $L_1$ and the hinge loss $L_2$. Specifically, we generate 100 observations from a mixture of two Gaussian distributions with equal probability: $N((-0.5, -0.5)^T, I_2)$ and $N((0.5, 0.5)^T, I_2)$ with $I_2$ an identity matrix of dimension two. In Figure 1, we plot the decision boundary $S(\boldsymbol{x}; \widehat{\boldsymbol{\theta}}_j)$ (in black) based on $\mathcal{D}_n$, and 100 perturbed decision boundaries $\{S(\boldsymbol{x}; \widehat{\boldsymbol{\theta}}_j^{*(1)}), \ldots, S(\boldsymbol{x}; \widehat{\boldsymbol{\theta}}_j^{*(100)})\}$ (in gray) for $j = 1, 2$; see Step 2 of *Algorithm 1*. Figure 1 reveals that the perturbed decision boundaries of the squared loss are more stable than those of the SVM given a small perturbation of the training sample. Hence, it is desirable to quantify the variability of the perturbed decision boundaries with respect to the original unperturbed decision boundary $S(\boldsymbol{x}; \widehat{\boldsymbol{\theta}}_j)$. This is a nontrivial task since the boundaries spread over a $d$-dimensional space, e.g., $d = 2$ in Figure 1. Therefore, we transform the data in such a way that the above variability can be fully measured in a single dimension. Specifically, we find a $d \times d$ transformation matrix $R_L$, which is orthogonal with determinant 1, such that the decision boundary based on the transformed data $\mathcal{D}_n^\dagger = \{(\boldsymbol{x}_i^\dagger, y_i), i = 1, \ldots, n\}$ with $\boldsymbol{x}_i^\dagger = R_L \boldsymbol{x}_i$ is parallel to the $\mathcal{X}_1, \ldots, \mathcal{X}_{d-1}$ axes; see the supplementary material S.3 for the calculation of $R_L$. The variability of the perturbed decision boundaries with respect to the original unperturbed decision boundary then reduces to the variability along the last axis $\mathcal{X}_d$. For illustration purposes, we next apply the above data-transformation idea to the SVM plotted in the middle plot of Figure 1. From the right plot in Figure 1, we observe that the variability of the transformed perturbed decision boundaries (in gray) with respect to the transformed unperturbed decision boundary (in black) now reduces to the variability along the $\mathcal{X}_2$ axis only. This is because the transformed unperturbed decision boundary is parallel to the $\mathcal{X}_1$ axis. Note that the choice of data transformation is not unique. For example, we could also transform the data such that the transformed unperturbed decision boundary is parallel to the $\mathcal{X}_2$ axis and then measure the variability along the $\mathcal{X}_1$ axis. Fortunately, the DBI measure we will introduce yields exactly the same value under any transformation, i.e., it is transformation invariant.

Figure 1: Two classes are shown in red circles and blue crosses. The black line is the decision boundary based on the original training sample, and the gray lines are 100 decision boundaries based on perturbed samples. The left (middle) panel corresponds to the least square loss (SVM). The perturbed decision boundaries of SVM after data transformation are shown on the right.

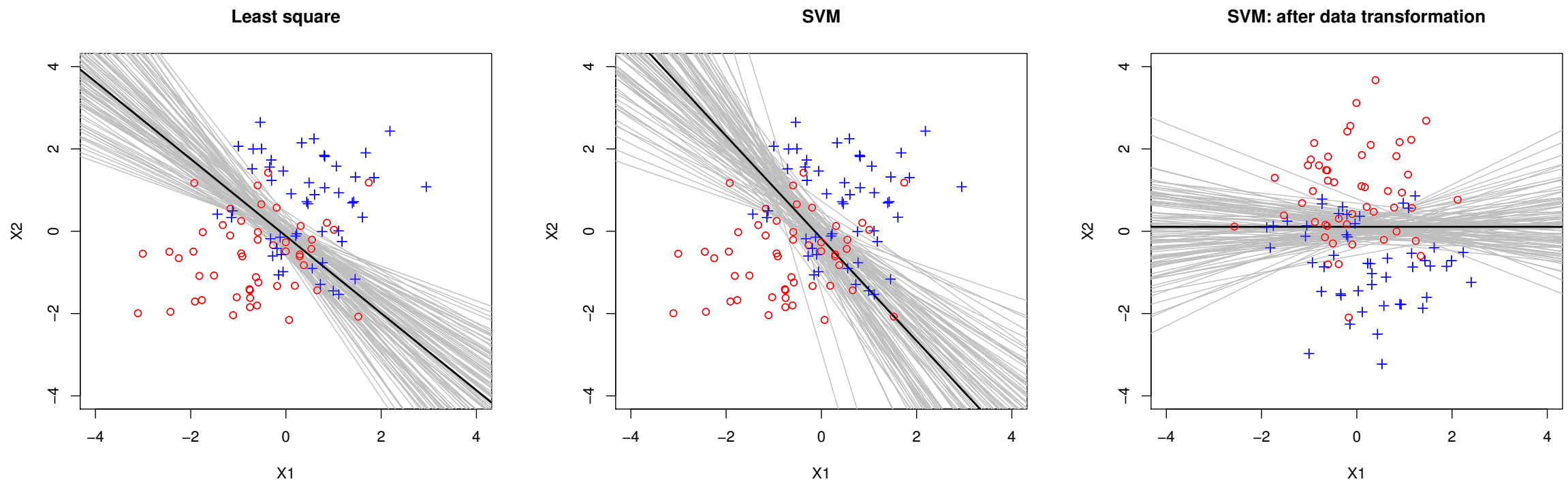

Now we are ready to define DBI. Given the loss function $L$, we define the coefficient estimator based on transformed data $\mathcal{D}_n^\dagger$ as $\widehat{\boldsymbol{\theta}}_L^\dagger$ and the coefficient estimator of its corresponding perturbed decision boundary as $\widehat{\boldsymbol{\theta}}_L^{\dagger *}$. We find the following relationship via the transformation matrix $R_L$:

$$\widehat{\boldsymbol{\theta}}_L \equiv \begin{pmatrix} \widehat{b}_L \\ \widehat{\boldsymbol{w}}_L \end{pmatrix} \Rightarrow \widehat{\boldsymbol{\theta}}_L^\dagger \equiv \begin{pmatrix} \widehat{b}_L \\ R_L \widehat{\boldsymbol{w}}_L \end{pmatrix} \text{ and } \widehat{\boldsymbol{\theta}}_L^* \equiv \begin{pmatrix} \widehat{b}_L^* \\ \widehat{\boldsymbol{w}}_L^* \end{pmatrix} \Rightarrow \widehat{\boldsymbol{\theta}}_L^{\dagger *} \equiv \begin{pmatrix} \widehat{b}_L^* \\ R_L \widehat{\boldsymbol{w}}_L^* \end{pmatrix},$$

which can be shown by replacing $\boldsymbol{x}_i$ with $R_L\boldsymbol{x}_i$ in (1) and (5) and using the property of $R_L$. Given $\widehat{\boldsymbol{\theta}}_L^{\dagger *} = (\widehat{b}_L^*, \widehat{w}_{L,1}^{\dagger *}, \ldots, \widehat{w}_{L,d}^{\dagger *})^T$, we define the $d$-th dimension of $S(\boldsymbol{X}; \widehat{\boldsymbol{\theta}}_L^{\dagger *})$ as

$$S_d := -\frac{\widehat{b}_L^{\dagger *}}{\widehat{w}_{L,d}^{\dagger *}} - \sum_{j=1}^{d-1} \frac{\widehat{w}_{L,j}^{\dagger *}}{\widehat{w}_{L,d}^{\dagger *}} X_j. \tag{7}$$

DBI is defined as the variability of the transformed perturbed decision boundary $S(\boldsymbol{X}; \widehat{\boldsymbol{\theta}}_L^{\dagger *})$ with respect to the transformed unperturbed decision boundary $S(\boldsymbol{X}; \widehat{\boldsymbol{\theta}}_L^{\dagger})$ along its $d$-th dimension.

**Definition 1** *The decision boundary instability (DBI) of $S(\boldsymbol{x}; \widehat{\boldsymbol{\theta}}_L)$ is defined to be*

$$DBI\Big(S(\boldsymbol{X}; \widehat{\boldsymbol{\theta}}_L)\Big) = E\left[Var\Big(S_d | \boldsymbol{X}_{(-d)}^\dagger\Big)\right], \tag{8}$$

*where $S_d$ is defined in (7) and $\boldsymbol{X}_{(-d)}^\dagger = (X_1^\dagger, \ldots, X_{d-1}^\dagger)^T$.*

**Remark 1** *The conditional variance* $Var(S_d|\boldsymbol{X}^{\dagger}_{(-d)})$ *in (8) captures the variability of the transformed perturbed decision boundary along the dth dimension based on a given sample. Note that, after data transformation, the transformed unperturbed decision boundary is parallel to the* $\mathcal{X}_1, \ldots, \mathcal{X}_{d-1}$ *axes. Therefore, this conditional variance precisely measures the variability of the perturbed decision boundary with respect to the unperturbed decision boundary conditioned on the given sample. The expectation in (8) then averages out the randomness in the sample.*

**Toy Example Continuation:** We next give an illustration of (8) via the 2-dimensional toy example shown in the right plot of Figure 1. For each sample, the conditional variance in (8) is estimated via the sample variability of the projected $X_2$ values on the perturbed decision boundary (in gray). Then the final DBI is estimated by averaging over all samples.

In Appendix A.1, we demonstrate an efficient way to simplify (8) by approximating the conditional variance via the weighted variance of $\widehat{\boldsymbol{\theta}}^{\dagger}_L$. The key idea is to connect the conditional variance of the $d$-th dimension of decision boundary with the variance of the coefficients of the corresponding decision function. Specifically, we show that

$$DBI\Big(S(\boldsymbol{X};\widehat{\boldsymbol{\theta}}_L)\Big) \approx (w^{\dagger}_{L,d})^{-2} E\left[\tilde{\boldsymbol{X}}^{\dagger T}_{(-d)} \left(n^{-1}\Sigma^{\dagger}_{0L,(-d)}\right) \tilde{\boldsymbol{X}}^{\dagger}_{(-d)}\right], \tag{9}$$

where $w^{\dagger}_{L,d}$ is the last entry of the transformed coefficient $\boldsymbol{\theta}^{\dagger}_{0L}$, and $n^{-1}\Sigma^{\dagger}_{0L,(-d)}$ is the asymptotic variance of the first $d$ dimensions of $\widehat{\boldsymbol{\theta}}^{\dagger}_L$. Therefore, DBI can be viewed as a proxy measure of the asymptotic variance of the decision function.

We next propose a plug-in estimate for the approximate version of DBI in (9). Direct estimation of DBI in (8) is possible, but it requires perturbing the transformed data. To reduce the computational cost, we can take advantage of our resampling results in Stage 1 based on the relationship between $\Sigma^{\dagger}_{0L}$ and $\Sigma_{0L}$. Specifically, we can estimate $\Sigma^{\dagger}_{0L}$ by

$$\widehat{\Sigma}^{\dagger}_L = \begin{pmatrix} \widehat{\Sigma}_b & \widehat{\Sigma}_{b,\boldsymbol{w}} R_L^T \\ R_L \widehat{\Sigma}_{\boldsymbol{w},b} & R_L \widehat{\Sigma}_{\boldsymbol{w}} R_L^T \end{pmatrix} \quad \text{given that} \quad \widehat{\Sigma}_L = \begin{pmatrix} \widehat{\Sigma}_b & \widehat{\Sigma}_{b,\boldsymbol{w}} \\ \widehat{\Sigma}_{\boldsymbol{w},b} & \widehat{\Sigma}_{\boldsymbol{w}} \end{pmatrix}, \tag{10}$$

where $\widehat{\Sigma}_L$ is the sample variance of $\widehat{\boldsymbol{\theta}}^{*}_L$ obtained from Stage 1 as a byproduct. Hence, combining (9) and (10), we propose the following DBI estimate:

$$\widehat{DBI}\Big(S(\boldsymbol{X};\widehat{\boldsymbol{\theta}}_L)\Big) = \frac{\sum_{i=1}^n \tilde{\boldsymbol{x}}^{\dagger T}_{i(-d)} \widehat{\Sigma}^{\dagger}_{L,(-d)} \tilde{\boldsymbol{x}}^{\dagger}_{i(-d)}}{(n\widehat{w}^{\dagger}_{L,d})^2}, \tag{11}$$

where $\widehat{w}^{\dagger}_{L,d}$ is the last entry of $\widehat{\boldsymbol{\theta}}^{\dagger}_L$, and $\widehat{\Sigma}^{\dagger}_{L,(-d)}$ is obtained by removing the last row and last column of $\widehat{\Sigma}^{\dagger}_L$ defined in (10). The DBI estimate in (11) is the one we will use in the numerical experiments.

## 3.3 Relationship of DBI with Other Variability Measures

In this subsection, we discuss the relationship of DBI with two alternative variability measures.

DBI may appear to be related to the asymptotic variance of the K-CV error, i.e., $E(\psi_1)^2$ in Theorem 1. However, we want to point out that these two quantities are quite different. For example, when data are nearly separable, reasonable perturbations to the data may only lead to a small variation in the K-CV error. On the other hand, small changes in the data (especially those support points near the decision boundary) may lead to a large variation in the decision boundary which implies a large DBI. This is mainly because DBI is conceptually different from the K-CV error. In Section 5, we provide concrete examples to show that these two variation measures generally lead to different choices of loss functions, and the loss function with the smallest DBI often corresponds to the classifier that is more accurate and stable.

Moreover, DBI shares similar spirit of the stability-oriented measure introduced in Bousquet and Elisseeff (2002). They defined theoretical stability measures for the purpose of deriving the generalization error bound. Their stability of a classification algorithm is defined as the maximal difference of the decision functions trained from the original dataset and the leave-one-out dataset. Their stability measure mainly focuses on the variability of the decision function and hence suffers from the transformation variant issue since a scale transformation of the decision function coefficients will greatly affect the value of a decision function. On the other hand, our DBI focuses on the variability of the decision boundary and is transformation invariant.

In the experiments, we will compare our classifier selection algorithm with approaches using these two alternative variability measures. Our method achieves superior performance in both classification accuracy and stability.

### 3.4 Summary of Classifier Selection Algorithm

In this section, we summarize our two-stage classifier selection algorithm.
*Algorithm 2 (Two-Stage Classifier Selection Procedure)*:
Input: Training sample $\mathcal{D}_n$ and a collection of candidate loss functions $\{L_j : j \in J\}$.

- Step 1. Obtain the K-CV errors $\widehat{\mathcal{D}}_j$ for each $j \in J$, and let the minimal value be $\widehat{\mathcal{D}}_t$.

- Step 2. Apply *Algorithm 1* to establish the pairwise confidence interval for each GE difference $\Delta_{tj}$. Eliminate the loss $L_j$ if the corresponding confidence interval does not cover zero. Specifically, the set of potentially good classifiers is defined to be

  $$\Lambda = \Big\{ j \in J : \widehat{\Delta}_{tj} - n^{-1/2}\phi_{t,j;\alpha/2} \le 0 \Big\},$$

  where $\widehat{\Delta}_{tj}$ and $\phi_{t,j;\alpha/2}$ are defined in Step 3 of *Algorithm 1*.

- Step 3. Estimate DBI for each $L_j$ with $j \in \Lambda$ via (11). The optimal loss function is $L_{j^*}$ with

  $$j^* = \arg\min_{j\in\Lambda} \widehat{DBI}\Big(S(\boldsymbol{X}; \widehat{\boldsymbol{\theta}}_j)\Big). \tag{12}$$

In Step 2, we fix the confidence level $\alpha = 0.1$ since it provides a sufficient but not too stringent confidence level. Our experiment in Section 6.1 further shows that the set $\Lambda$ is quite stable against $\alpha$ within a reasonable range around 0.1. The optimal loss function $L_{j^*}$ selected in (12) is not necessarily unique. However, according to our experiments, multiple optimal loss functions are quite uncommon. Although in principle we can also perform an additional significance test for DBI in Step 3, the related computational cost is high given that DBI is already a second-moment measure. Hence, we choose not to include this test in our algorithm.

# 4 Selection Consistency

This section investigates the selection consistency of our algorithm by showing that the selected classifier achieves the minimal GE and minimal DBI asymptotically. To simplify the presentation, we establish our selection consistency via the large-margin unified machines (LUM, Liu et al., 2011); the extension to other large-margin classifiers is straightforward.

The LUM offers a platform unifying various large margin classifiers ranging from soft ones to hard ones. A soft classifier estimates the class conditional probabilities explicitly and makes the class prediction via the largest estimated probability, while a hard classifier directly estimates the classification boundary without a class-probability estimation (Wahba, 2002). The class of LUM loss functions can be written as

$$L_\gamma(u) = \begin{cases} 1-u & \text{if } u < \gamma \\ (1-\gamma)^2(\frac{1}{u-2\gamma+1}) & \text{if } u \geq \gamma, \end{cases} \tag{13}$$

where the index parameter $\gamma \in [0,1]$. As shown by Liu et al. (2011), when $\gamma = 1$ the LUM loss reduces to the hinge loss of SVM, which is a typical example of hard classification; when $\gamma = 0.5$ the LUM loss is equivalent to the DWD classifier, which can be viewed as a classifier that is between hard and soft; and when $\gamma = 0$ the LUM loss becomes a soft classifier that has an interesting connection with the logistic loss. Therefore, the LUM framework approximates many of the soft and hard classifiers in the literature. Figure 2 displays LUM loss functions for various values of $\gamma$ and compares them with some commonly used loss functions.

Figure 2: Plots of least square, exponential, logistic, and LUM loss functions with $\gamma = 0, 0.5, 1$.

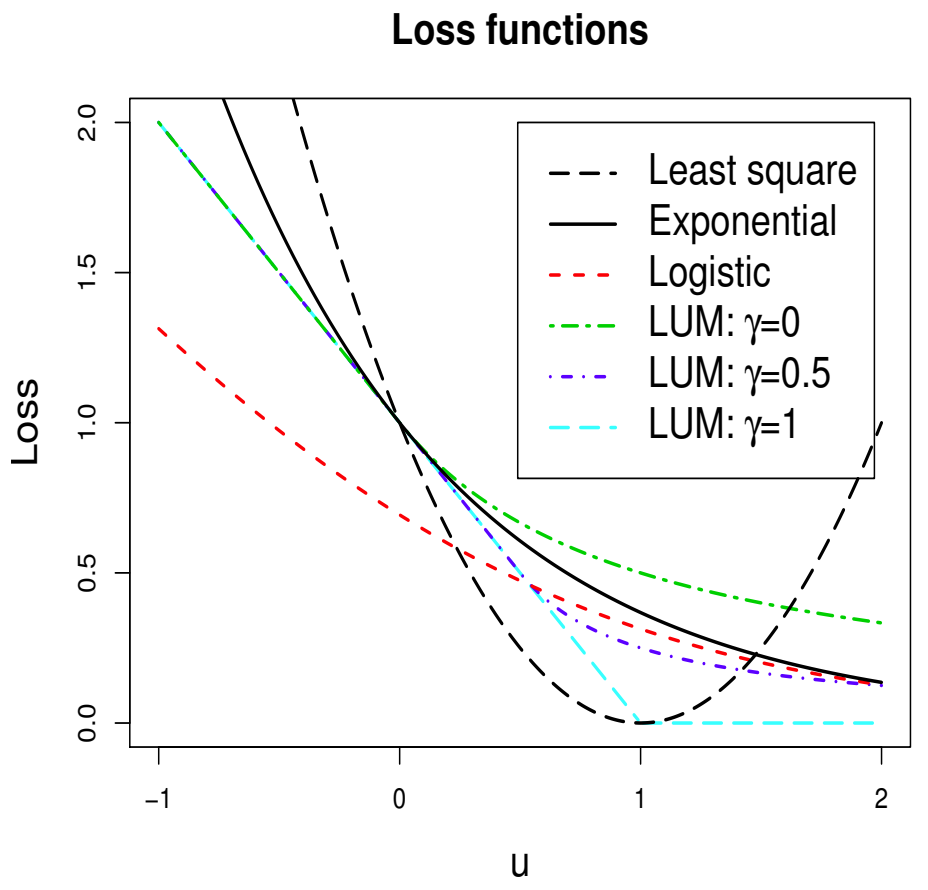

In the LUM framework, we denote the true risk as $\mathcal{R}_\gamma(\boldsymbol{\theta}) = E[L_\gamma(yf(\boldsymbol{x};\boldsymbol{\theta}))]$, the true parameter as $\boldsymbol{\theta}_{0\gamma} = \arg\min_{\boldsymbol{\theta}} \mathcal{R}_\gamma(\boldsymbol{\theta})$, the GE as $D_{0\gamma}$, the empirical generalization error as $\widehat{D}_\gamma$, and the K-CV error as $\widehat{\mathcal{D}}_\gamma$. In practice, given data $\mathcal{D}_n$, LUM solves

$$\widehat{\boldsymbol{\theta}}_\gamma = \arg\min_{b,\boldsymbol{w}} \left\{ \frac{1}{n} \sum_{i=1}^{n} L_\gamma\Big(y_i(\boldsymbol{w}^T \boldsymbol{x}_i + b)\Big) + \frac{\lambda_n \boldsymbol{w}^T \boldsymbol{w}}{2} \right\}. \tag{14}$$

In Corollary 1 and Corollary 2 provided in Section S.4 of the online supplement, we establish the asymptotic normality of $\widehat{\boldsymbol{\theta}}_\gamma$ and $\widehat{\mathcal{D}}_\gamma$, respectively. These preliminary results are used to develop the following selection consistency of our two-stage classifier selection algorithm.

For the LUM class, we define the set of potentially good classifiers as

$$\widehat{\Lambda}_0 = \Big\{ \gamma \in [0,1] : \widehat{\mathcal{D}}_\gamma \le \widehat{\mathcal{D}}_{\widehat{\gamma}_0^*} + n^{-1/2} \phi_{\gamma,\widehat{\gamma}_0^*;\alpha/2} \Big\}, \tag{15}$$

where $\widehat{\gamma}_0^* = \arg\min_{\gamma\in[0,1]} \widehat{\mathcal{D}}_\gamma$, based on $\mathcal{D}_n$. Its population version is thus defined as those classifiers achieving the minimal GE, denoted

$$\Lambda_0 = \Big\{ \gamma \in [0,1] : D_{0\gamma} = D_{0\gamma_0^*} \Big\}, \tag{16}$$

where $\gamma_0^* = \arg\min_{\gamma\in[0,1]} D_{0\gamma}$. To show the selection consistency, we require an additional assumption on the Hessian matrix $H(\boldsymbol{\theta}_{0\gamma})$ defined in Corollary 1 in Section S.4 of the online supplement:

(B1) The smallest eigenvalue of the true Hessian matrix $\lambda_{\min}(H(\boldsymbol{\theta}_{0\gamma})) \ge c_1$, and the largest eigenvalue of the true Hessian matrix $\lambda_{\max}(H(\boldsymbol{\theta}_{0\gamma})) \le c_2$, where the positive constants $c_1, c_2$ do not depend on $\gamma$.

As seen in the proof of Corollary 1, the true Hessian matrix $H(\boldsymbol{\theta}_{0\gamma})$ is positive definite for any fixed $\gamma \in [0,1]$. Therefore, Assumption (B1) is slightly stronger in the uniform sense. It is required to guarantee the uniform convergence results, i.e., ($S.15$) and ($S.17$), in Section S.7 of the online supplement.

Our Lemma 1 first ensures that the minimum K-CV error converges to the minimum GE at a root-n rate.

**Lemma 1** *Suppose that Assumptions (L1), (B1), and Assumption (A1) in Section S.4 of the online supplement hold. We have, if* $\lambda_n = o(n^{-1/2})$,

$$\Big| \widehat{\mathcal{D}}_{\widehat{\gamma}_0^*} - D_{0\gamma_0^*} \Big| = O_P(n^{-1/2}). \tag{17}$$

In the second stage, we denote the index of the selected optimal classifier as

$$\widehat{\gamma}_0 = \arg\min_{\gamma\in\widehat{\Lambda}_0} \widehat{DBI}\Big(S(\boldsymbol{X}; \widehat{\boldsymbol{\theta}}_\gamma)\Big), \tag{18}$$

and its population version as

$$\gamma_0 = \arg\min_{\gamma\in\Lambda_0} DBI\Big(S(\boldsymbol{X}; \widehat{\boldsymbol{\theta}}_\gamma)\Big). \tag{19}$$

**Theorem 3** *Suppose that the assumptions in Lemma 1 hold. We have, as $N \to \infty$,*

$$\left|\widehat{DBI}\Big(S(\boldsymbol{X}; \widehat{\boldsymbol{\theta}}_{\widehat{\gamma}_0})\Big) - DBI\Big(S(\boldsymbol{X}; \widehat{\boldsymbol{\theta}}_{\gamma_0})\Big)\right| = o_P(n^{-1}). \tag{20}$$

*Recall that $N$ is the number of resamplings defined in Step 2 of Algorithm 1.*

Theorem 3 implies that the estimated DBI of the selected classifier converges to the DBI of the true optimal classifier, which has the smallest DBI. Therefore, the proposed two-stage algorithm is able to select the classifier with the minimal DBI among those classifiers having the minimal GE. In summary, we have shown that the selected optimal classifier has achieved the minimal GE and the minimal DBI asymptotically.

# 5 Experiments

In this section, we first demonstrate the DBI estimation procedure introduced in Section 3.2, and then illustrate the applicability of our classifier selection method in various simulated and real examples. In all experiments, we compare our selection procedure, denoted as "cv+dbi", with two alternative methods: 1) "cv+varcv" which is the two-stage approach selecting the loss with the minimal variance of the K-CV error in Stage 2, and 2) "cv+be" which is the two-stage approach selecting the loss with the minimal classification stability defined in Bousquet and Elisseeff (2002) in Stage 2. Stage 1 of each alternative approach is the same as ours. We consider six large-margin classifier candidates: least squares loss, exponential loss, logistic loss, and LUM with $\gamma = 0, 0.5, 1$. Recall that LUM with $\gamma = 0.5$ ($\gamma = 1$) is equivalent to DWD (SVM). In all the large-margin classifiers, the tuning parameter $\lambda_n$ is selected via cross-validation.

## 5.1 Illustration

This subsection demonstrates the DBI estimation procedure and checks the sensitivity of the confidence level $\alpha$ in Algorithm 2.

We generated labels $y \in \{-1, 1\}$ with equal probability. Given $Y = y$, the predictor vector $(x_1, x_2)$ was generated from a bivariate normal $N((\mu y, \mu y)^T, I_2)$ with the signal level $\mu = 0.8$.

We first illustrate the DBI estimation procedure in Section 3.2 by comparing the estimated DBIs with the true DBIs for various sample sizes. We varied the sample size $n$ among 50, 100, 200, 500, and 1000. The classifier with the least squares loss was investigated due to its simplicity. Simple algebra implied that the true parameter $\boldsymbol{\theta}_{0L} = (0, 0.351, 0.351)$ and the transformed parameter $\boldsymbol{\theta}_{0L}^{\dagger} = (0, 0, 0.429)$. Furthermore, the covariance matrix $\Sigma_{0L}$ and the transformed covariance matrix $\Sigma_{0L}^{\dagger}$ were computed as

$$\Sigma_{0L} = \begin{pmatrix} 0.439 & 0 & 0 \\ 0 & 0.268 & -0.170 \\ 0 & -0.170 & 0.268 \end{pmatrix} \quad \text{and} \quad \Sigma_{0L}^{\dagger} = \begin{pmatrix} 0.439 & 0 & 0 \\ 0 & 0.439 & 0 \\ 0 & 0 & 0.098 \end{pmatrix},$$

given the transformation matrix

$$R_L = \begin{pmatrix} -\frac{\sqrt{2}}{2} & \frac{\sqrt{2}}{2} \\ \frac{\sqrt{2}}{2} & \frac{\sqrt{2}}{2} \end{pmatrix}.$$

Finally, plugging all these terms into (9) led to

$$DBI\Big(S(\boldsymbol{X};\widehat{\boldsymbol{\theta}}_L)\Big) \approx \frac{3.563}{n}. \tag{21}$$

Figure 3 compares the estimated DBIs in (11) with the true DBIs in (21). Clearly, they match very well for various sample sizes and their difference vanishes as the sample size increases. This experiment empirically validates the estimation procedure in Section 3.2.

Figure 3: Comparison of true and estimated DBIs in Example 6.1. The true DBIs are denoted as red triangles and the estimated DBIs from replicated experiments are illustrated by box plots.

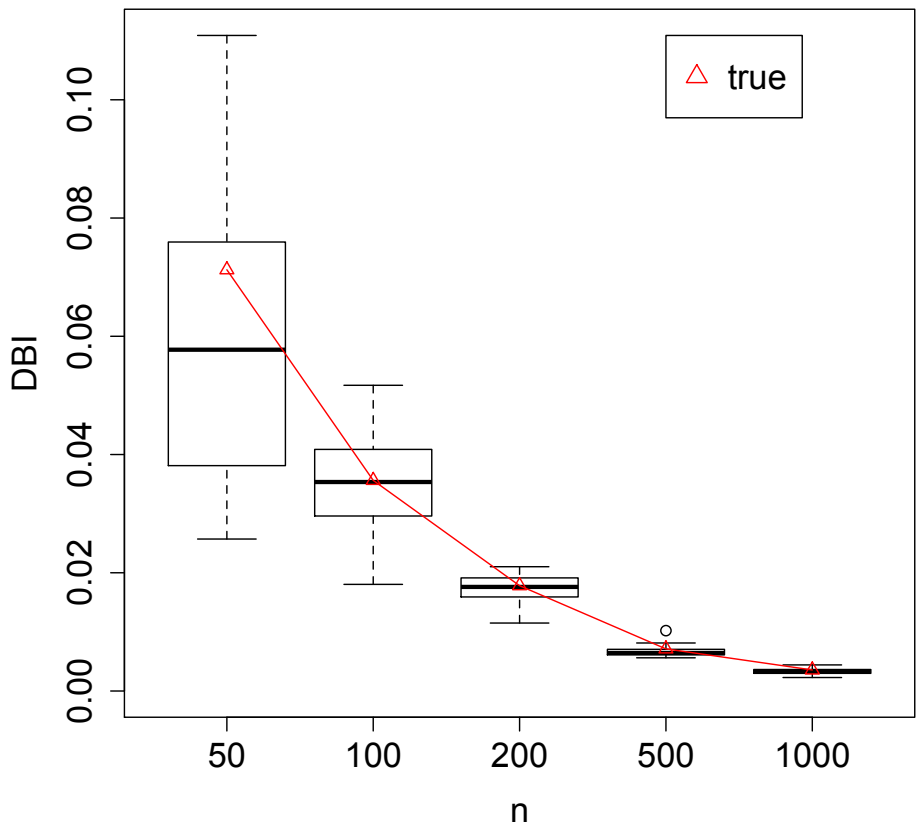

In order to show the sensitivity of the confidence level $\alpha$ to the set $\Lambda$ in Algorithm 2, we randomly selected one replication and display the proportion of potentially good classifiers over all six classifiers. Note that as $\alpha$ increases, the confidence interval for the difference of GEs will be narrower, and hence the size of $\Lambda$ will be smaller. Therefore, the change of the proportion reflects exactly the change of $\Lambda$ since $\Lambda$ is monotone with respect to $\alpha$. For each $\alpha \in \{l/100;\ l = 0, \ldots, 50\}$, we computed the proportion of potentially good classifiers and observed that the proportion was stable in a reasonable large range around 0.1.

## 5.2 Simulations

In this section, we illustrate the superior performance of our method using four simulated examples. These simulations were previously studied by Liu et al. (2011). In all of the simulations, the size of training data sets was 100 and that of testing data sets was 1000. All the procedures were repeated 100 times and the averaged test errors and averaged test DBIs of the selected classifier were reported.

**Simulation 1**: Two predictors were uniformly generated over $\{(x_1, x_2) : x_1^2+x_2^2 \leq 1\}$. The class label $y$ was 1 when $x_2 \geq 0$ and $-1$ otherwise. We generated 100 samples and then contaminated the data by randomly flipping the labels of 15% of the instances.

**Simulation 2**: The setting was the same as Simulation 1 except that we contaminated the data by randomly flipping the labels of 25% of the instances.

**Simulation 3**: The setting was the same as Simulation 1 except that we contaminated the data by randomly flipping the labels of 80% of the instances whose $|x_2| \geq 0.7$.

**Simulation 4**: Two predictors were uniformly generated over $\{(x_1, x_2) : |x_1| + |x_2| \leq 2\}$. Conditionally on $X_1 = x_1$ and $X_2 = x_2$, the class label $y$ took 1 with probability $e^{3(x_1+x_2)}/(1 + e^{3(x_1+x_2)})$ and $-1$ otherwise.

We first demonstrate the mechanism of our proposed method for one repetition of Simulation 1. As shown in the upper left plot of Figure 4, exponential loss and LUMs with $\gamma = 0.5$ or 1 are potentially good classifiers in Stage 1; they happen to have the same K-CV error. Their corresponding DBIs are compared in the second stage. As shown in the upper right plot of Figure 4, LUM with $\gamma = 0.5$ gives the minimal DBI and is selected as the final classifier. In this example, although exponential loss also gives the minimal K-CV error, its decision boundary is unstable compared to that of LUM with $\gamma = 0.5$. This shows that the K-CV estimate itself is not sufficient for classifier comparison, since it ignores the variation in the classifier. To show that our DBI estimation is reasonable, we display the perturbed decision boundaries for these three potentially good classifiers on the bottom of Figure 4. The relationship among their instabilities is precisely captured by our DBI estimate: compared with the exponential loss and LUM with $\gamma = 1$, LUM with $\gamma = 0.5$ is more stable.

Figure 4: The K-CV error, the DBI estimate, and the perturbed decision boundaries in Simulation 1 with flipping rate 15%. The minimal K-CV error and minimal DBI estimate are indicated with red triangles. The labels Ls, Exp, Logit, LUM0, LUM0.5, and LUM1 refer to least squares loss, exponential loss, logistic loss, and LUM loss with index $\gamma = 0, 0.5, 1$, respectively.

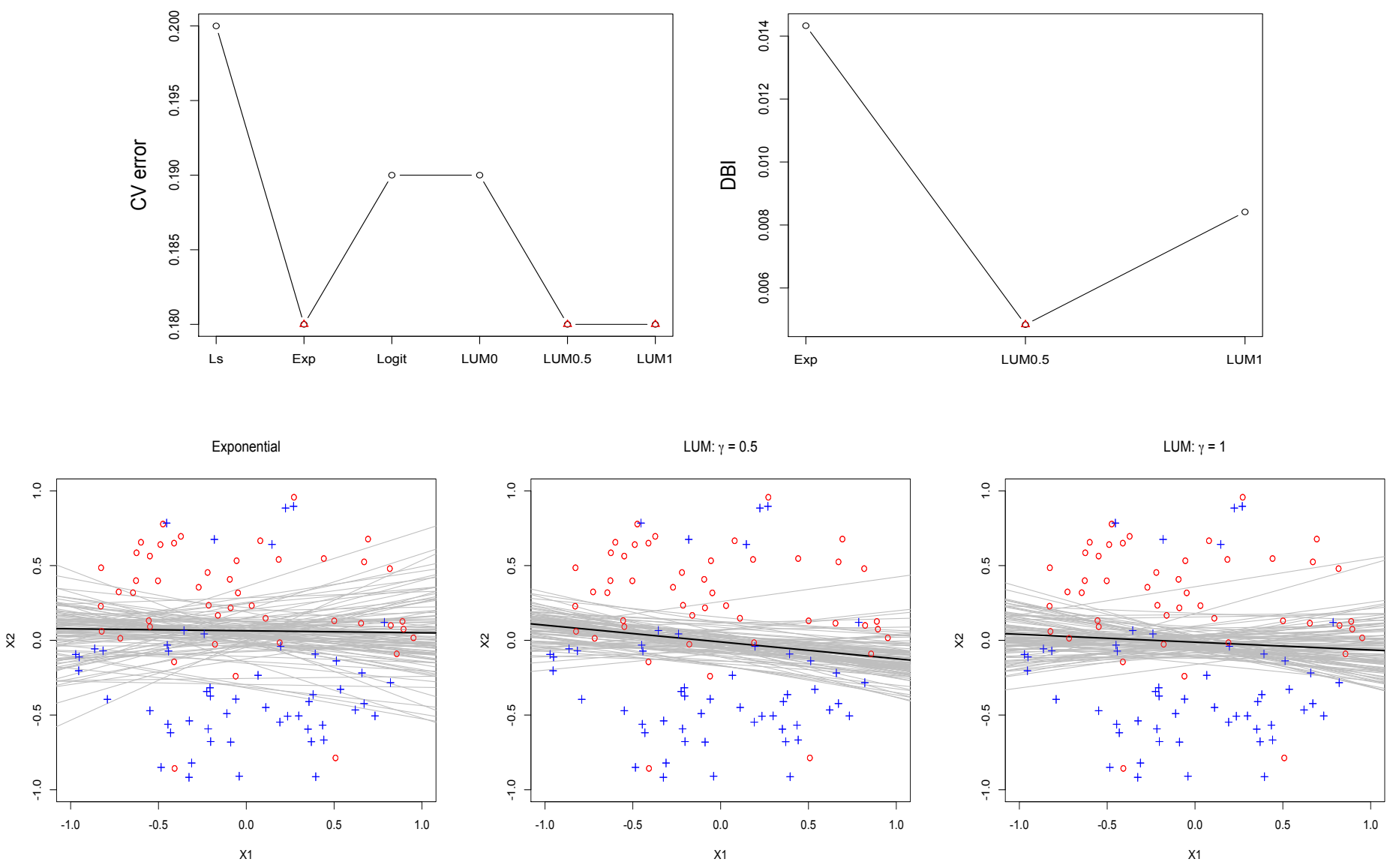

We report the averaged test errors and averaged test DBIs of the classifier selected from our method as well as two alternative approaches, see Table 1. In all four simulated examples, our "cv+dbi" achieves the smallest test errors, while the difference among test errors of all algorithms is generally not significant. This phenomenon of indistinguishable test errors agrees with the fact that all methods are the same during the first stage and those left from Stage 1 are all potentially good in terms of classification accuracy. However, our "cv+dbi" is able to choose the classifiers with minimal test DBIs in all simulations and the improvements over other algorithms are significant. Overall, our method is able to choose the classifier with outstanding performance in both classification accuracy and stability.

Table 1: The averaged test errors and averaged test DBIs (multiplied by 100) of all methods: "cv+varcv" is the two-stage approach which selects the loss with the minimal variance of the K-CV error in Stage 2; "cv+be" is the two-stage approach which in Stage 2 selects the loss with the minimal classification stability defined in Bousquet and Elisseeff (2002); "cv+dbi" is our method. The smallest value in each case is given in bold. Standard errors are given in subscript.

| Simulations | | cv+varcv | cv+be | cv+dbi |
|---|---|---|---|---|
| Sim 1 | Error | $0.191_{0.002}$ | $0.194_{0.002}$ | $\mathbf{0.190}_{0.002}$ |
| | DBI | $0.139_{0.043}$ | $0.135_{0.019}$ | $\mathbf{0.081}_{0.002}$ |
| Sim 2 | Error | $0.296_{0.002}$ | $0.303_{0.003}$ | $\mathbf{0.295}_{0.002}$ |
| | DBI | $0.291_{0.044}$ | $0.318_{0.036}$ | $\mathbf{0.229}_{0.012}$ |
| Sim 3 | Error | $0.218_{0.006}$ | $0.234_{0.006}$ | $\mathbf{0.209}_{0.004}$ |
| | DBI | $0.124_{0.008}$ | $0.291_{0.037}$ | $\mathbf{0.107}_{0.003}$ |
| Sim 4 | Error | $0.120_{0.001}$ | $0.121_{0.001}$ | $\mathbf{0.119}_{0.001}$ |
| | DBI | $0.884_{0.207}$ | $0.414_{0.106}$ | $\mathbf{0.235}_{0.038}$ |

### 5.3 Real Examples

In this subsection, we compare our method with the alternatives on three real datasets in the UCI Machine Learning Repository (Frank and Asuncion, 2010).

The first data set is the liver disorders data set (*liver*) which consists of 345 samples with 6 variables of blood test measurements. The class label splits the data into 2 classes with sizes 145 and 200. The second data set is the breast cancer data set (*breast*) which consists of 683 samples after removing missing values (Wolberg and Mangasarian, 1990). Each sample has 10 experimental measurement variables and one binary class label indicating whether the sample is benign or malignant. These 683 samples arrived periodically as Dr. Wolberg reported his clinical cases. In total, there are 8 groups of samples which reflect the chronological order of the data. It is expected that a good classification procedure should generate a classifier that is stable across these groups of samples. The third data set is the credit approval data set (*credit*) which consists of 690 samples with 15 features, among which 307 samples have a positive class label and the rest 383 samples have a negative class label.

For each dataset, we randomly split the data into 2/3 training samples and 1/3 testing samples, and reported the averaged test errors and averaged test DBIs based on all classifier selection algorithms over 50 replications, see Table 2. Compared with the alternatives, our "cv+dbi" method obtains significant improvements in DBIs and simultaneously attains satisficatory test errors that are minimal or statistically indistinguishable to the minimal one. This indicates that the proposed method could serve as a practical tool for selecting an accurate and stable classifier.

Table 2: The averaged test errors and averaged test DBIs of all methods in real example. The smallest value in each case is given in bold. Standard errors are given in subscript.

| Data | | cv+varcv | cv+be | cv+dbi |
|---|---|---|---|---|
| Liver | Error | $0.331_{0.006}$ | $0.335_{0.006}$ | $\mathbf{0.327}_{0.006}$ |
| | DBI | $0.140_{0.013}$ | $0.157_{0.024}$ | $\mathbf{0.113}_{0.012}$ |
| Breast | Error | $\mathbf{0.038}_{0.002}$ | $\mathbf{0.038}_{0.002}$ | $\mathbf{0.038}_{0.002}$ |
| | DBI | $0.388_{0.066}$ | $0.152_{0.028}$ | $\mathbf{0.124}_{0.023}$ |
| Credit | Error | $\mathbf{0.135}_{0.004}$ | $0.138_{0.004}$ | $0.136_{0.004}$ |
| | DBI | $0.229_{0.101}$ | $0.157_{0.042}$ | $\mathbf{0.112}_{0.023}$ |

# 6 Discussion

This paper proposes a two-stage classifier selection procedure based on GE and DBI. It selects the classifier with the most stable decision boundary among those classifiers with relatively small estimated GEs. The concept of DBI is quite general, and its extension to a broader framework, e.g., multi-category classification (Shen and Wang, 2007; Zhang and Liu, 2013) or high-dimensional classification (Fan et al., 2012), is conceptually simple. In particular, in the multi-category classification, we suggest to use the one-versus-all idea (Rifkin and Klautau, 2004) to extend our DBI measure. For $K$ classes, we compute $\text{DBI}_k$ as the DBI between the $k$-th class and the rest $K-1$ classes, and then average the DBIs to obtain the final DBI as $K^{-1}\sum_{k=1}^{K}\text{DBI}_k$. When $K = 2$, it reduces to our original DBI.

It is worth noting that the extension to the nonlinear classifiers is also feasible. We will give detailed discussions on the nonlinear extension in Appendix A.2. In short, in Stage 1, the asymptotic normality of the nonlinear K-CV error is still valid due to Hable (2012); in Stage 2, measuring the instability of the nonlinear decision boundaries is possible by mapping the nonlinear decision boundaries to a higher dimensional space where the projected decision boundaries are linear.

# Supplementary Materials

In the online supplement, we provide technical proofs of all lemmas, corollaries and theorems, discuss the calculation of the transformation matrix, and provide a notation table.

# Acknowledgment

The authors would like to thank the Co-Editor Professor Hsin-Cheng Huang, Associate Editor and two referees for their constructive comments and suggestions, which have led to a significantly improved paper. We would like to thank SAMSI for warm hosting when part of the work was done. Will Wei Sun would like to thank Yongheng Zhang for discussions on the transformation matrix.

# Appendix: Technical Details

In the Appendix, we discuss an efficient approximation of DBI, and propose a nonlinear extension of our two-stage classifier selection algorithm.

## A.1. Approximating DBI via (9)

We propose an approximate version of DBI, i.e., (9), which can be easily estimated in practice. According to (8), we can calculate $DBI(S(\boldsymbol{X};\widehat{\boldsymbol{\theta}}_L))$ as

$$E\left[\tilde{\boldsymbol{X}}_{(-d)}^{\dagger T} Var\left(\widehat{\boldsymbol{\eta}}_L^{\dagger *}|\boldsymbol{X}_{(-d)}^{\dagger}\right)\tilde{\boldsymbol{X}}_{(-d)}^{\dagger}\right], \tag{A.1}$$

where $\tilde{\boldsymbol{X}}_{(-d)}^{\dagger} = (1, \boldsymbol{X}_{(-d)}^{\dagger T})^T$ and $\widehat{\boldsymbol{\eta}}_L^{\dagger *} = \left(-\widehat{b}_L^{\dagger *}/\widehat{w}_{L,d}^{\dagger *}, -\widehat{w}_{L,1}^{\dagger *}/\widehat{w}_{L,d}^{\dagger *} \ldots, -\widehat{w}_{L,d-1}^{\dagger *}/\widehat{w}_{L,d}^{\dagger *}\right)$. To further simplify (A.1), we need the following theorem as an intermediate step.

**Theorem 4** *Suppose that Assumptions (L1)–(L5) hold and $\lambda_n = o(n^{-1/2})$. We have, as $n \to \infty$,*

$$\sqrt{n}(\widehat{\boldsymbol{\theta}}_L - \boldsymbol{\theta}_{0L}) \xrightarrow{d} N(0, \Sigma_{0L}), \tag{A.2}$$

$$\sqrt{n}(\widehat{\boldsymbol{\theta}}_L^{*} - \widehat{\boldsymbol{\theta}}_L) \overset{d}{\Longrightarrow} N(0, \Sigma_{0L}) \quad \textit{conditional on } \mathcal{D}_n, \tag{A.3}$$

*where $\Sigma_{0L} = H(\boldsymbol{\theta}_{0L})^{-1}G(\boldsymbol{\theta}_{0L})H(\boldsymbol{\theta}_{0L})^{-1}$. After data transformation, we have, as $n \to \infty$,*

$$\sqrt{n}(\widehat{\boldsymbol{\theta}}_L^{\dagger} - \boldsymbol{\theta}_{0L}^{\dagger}) \xrightarrow{d} N(0, \Sigma_{0L}^{\dagger}), \tag{A.4}$$

$$\sqrt{n}(\widehat{\boldsymbol{\theta}}_L^{\dagger *} - \widehat{\boldsymbol{\theta}}_L^{\dagger}) \overset{d}{\Longrightarrow} N(0, \Sigma_{0L}^{\dagger}) \quad \textit{conditional on } \mathcal{D}_n^{\dagger}, \tag{A.5}$$

*where $\boldsymbol{\theta}_{0L}^{\dagger} = (b_{0L}, \boldsymbol{w}_{0L}^T R_L^T)^T$ and*

$$\Sigma_{0L}^{\dagger} = \begin{pmatrix} \Sigma_b & \Sigma_{b,\boldsymbol{w}} R_L^T \\ R_L \Sigma_{\boldsymbol{w},b} & R_L \Sigma_{\boldsymbol{w}} R_L^T \end{pmatrix} \quad \textit{if we partition } \Sigma_{0L} \textit{ as } \begin{pmatrix} \Sigma_b & \Sigma_{b,\boldsymbol{w}} \\ \Sigma_{\boldsymbol{w},b} & \Sigma_{\boldsymbol{w}} \end{pmatrix}.$$

We omit the proof of Theorem 4 since (*A.2*) and (*A.3*) directly follow from (S.1) and Appendix D in Jiang et al. (2008), and (A.4) and (A.5) follow from the Delta method.

Let $\widehat{\boldsymbol{\eta}}_L^{\dagger} = \left(-\widehat{b}_L^{\dagger}/\widehat{w}_{L,d}^{\dagger}, -\widehat{w}_{L,1}^{\dagger}/\widehat{w}_{L,d}^{\dagger} \ldots, -\widehat{w}_{L,d-1}^{\dagger}/\widehat{w}_{L,d}^{\dagger}\right)$. According to (A.4) and (A.5), we know that $Var(\widehat{\boldsymbol{\eta}}_L^{\dagger *}|\boldsymbol{X}_{(-d)}^{\dagger})$ is a consistent estimate of $Var(\widehat{\boldsymbol{\eta}}_L^{\dagger})$ because $\widehat{\boldsymbol{\eta}}_L^{\dagger *}$ and $\widehat{\boldsymbol{\eta}}_L^{\dagger}$ can be written as the same function of $\widehat{\boldsymbol{\theta}}_L^{\dagger *}$ and $\widehat{\boldsymbol{\theta}}_L^{\dagger}$, respectively. Hence, we claim that

$$DBI\left(S(\boldsymbol{X};\widehat{\boldsymbol{\theta}}_L)\right) \approx E\left(\tilde{\boldsymbol{X}}_{(-d)}^{\dagger T} Var(\widehat{\boldsymbol{\eta}}_L^{\dagger})\tilde{\boldsymbol{X}}_{(-d)}^{\dagger}\right).$$

Furthermore, we can approximate $Var(\widehat{\boldsymbol{\eta}}_L^{\dagger})$ by $(w_{L,d}^{\dagger})^{-2}[n^{-1}\Sigma_{0L,(-d)}^{\dagger}]$, where $n^{-1}\Sigma_{0L,(-d)}^{\dagger}$ is the asymptotic variance of the first $d$ dimensions of $\widehat{\boldsymbol{\theta}}_L^{\dagger}$, since $\widehat{w}_{L,d}^{\dagger}$ asymptotically follows the normal distribution with mean $w_{L,d}^{\dagger}$ and variance converging to 0 as $n$ grows (Hinkley, 1969). Finally, we can get the desirable approximation (9) for DBI. ∎

## A.2. Nonlinear Extension

The extension of our two-stage algorithm to nonlinear classifiers contains two aspects: (1) asymptotic normality of the K-CV error in Stage 1; (2) the application of DBI in Stage 2. The former is still valid due to Hable (2012), and the latter is feasible by mapping the nonlinear decision

**Extension of Stage 1**: We first modify several key concepts. The loss $L : \mathcal{X} \times \mathcal{Y} \times \mathbb{R} \to [0, \infty)$ is convex if it is convex in its third argument for every $(x, y) \in \mathcal{X} \times \mathcal{Y}$. A reproducing kernel Hilbert space (RKHS) H is a space of functions $f : \mathcal{X} \to \mathbb{R}$ which is generated by a kernel $k : \mathcal{X} \times \mathcal{X} \to \mathbb{R}$. Here the kernel $k$ could be a linear kernel, a Gaussian RBF kernel, or a polynomial kernel.

Given i.i.d training samples $\mathcal{D}_n = \{(\boldsymbol{x}_i, y_i); i = 1, \dots, n\}$ drawn from $P = (X, Y)$, the empirical function $f_{L,\mathcal{D}_n,\lambda_n}$ solves

$$\min_{f \in \mathcal{H}} \frac{1}{n} \sum_{i=1}^{n} L(x_i, y_i, f(x_i)) + \lambda_n \|f\|_{\mathcal{H}}^2.$$

In the nonparametric case, the optimization problem of minimizing population risk is ill-posed because a solution is not necessarily unique, and small changes in $P$ may have large effects on the solution. Therefore it is common to impose a bound on the complexity of the predictor and estimate a smoother approximation to the population version (Hable, 2012). For a fixed $\lambda_0 \in (0, \infty)$, we denote $f_{L,P,\lambda_0}$ as the population function which solves

$$\min_{f \in \mathcal{H}} \int L(x, y, f(x)) P(d(x, y)) + \lambda_0 \|f\|_{\mathcal{H}}^2.$$

The following conditions are assumed in Hable (2012) to prove the asymptotic normality of the estimated kernel decision function.

(N1) The loss $L$ is a convex, P-square-integrable Nemitski loss function of order $p \in [1, \infty)$. That is, there is a P-square-integrable function $b : \mathcal{X} \times \mathcal{Y} \to R$ such that $|L(x, y, t)| \leq b(x, y) + |t|^p$ for every $(x, y, t) \in \mathcal{X} \times \mathcal{Y} \times \mathbb{R}$.

(N2) The partial derivatives $L^{'}(x, y, t) := \frac{\partial L}{\partial t}(x, y, t)$ and $L^{''}(x, y, t) := \frac{\partial^2 L}{\partial^2 t}(x, y, t)$ exist for every $(x, y, t) \in \mathcal{X} \times \mathcal{Y} \times \mathbb{R}$ and are continuous.

(N3) For every $a \in (0, \infty)$, there is $b_a^{'} \in L_2(P)$ and $b_a^{''} \in [0, \infty)$ such that, for every $(x, y) \in \mathcal{X} \times \mathcal{Y}$, $\sup_{t \in [-a,a]} |L^{'}(x, y, t)| \leq b_a^{'}(x, y)$ and $\sup_{t \in [-a,a]} |L^{''}(x, y, t)| \leq b_a^{''}$.

**Proposition 1** *(Theorem 3.1, Hable (2012)) Under Assumptions (N1)-(N3) and* $\lambda_n = \lambda_0 + o(n^{-1/2})$, *for every* $\lambda_0 \in (0, \infty)$, *there is a tight, Borel-measurable Gaussian process* $\mathbb{H} : \Omega \to H$ *such that* $\sqrt{n}\Big(f_{L,\mathcal{D}_n,\lambda_n} - f_{L,P,\lambda_0}\Big) \to \mathbb{H}$.

**Remark 2** *Among the loss functions considered in this paper, the least squares, exponential, and logistic losses all satisfy the assumptions (N1)-(N3), while the LUM loss is not differentiable and does not satisfy Assumption (N2). However, Hable (2012) showed that any Lipschitz-continuous loss function (e.g. LUM loss) can always be modified as a differentiable* $\epsilon-$*version of the loss function such that the assumptions (N1)-(N3) are satisfied; see Remark 3.5 in Hable (2012).*

In the nonlinear case, the GE $D_{0L}$ and the K-CV error $\widehat{\mathcal{D}}_L$ are modified accordingly. The asymptotic normality of $\mathcal{W}_L = \sqrt{n}(\widehat{\mathcal{D}}_L - D_{0L})$ follows from Proposition 1, Corollary 3.3 in Hable (2014), and a slight modification of the proof of our Theorem 1. Then a perturbation-based resampling approach can be constructed analogously to Algorithm 1.

**Extension of Stage 2**: The concept of DBI is defined for linear decision boundaries. In order to measure the instability of nonlinear decision boundaries, we can map the nonlinear decision boundaries to a higher dimensional space where the projected decision boundaries are linear.

Figure A1: The nonlinear perturbed decision boundaries for the least squares loss (left) and SVM (right) in the bivariate normal example with unequal variances.

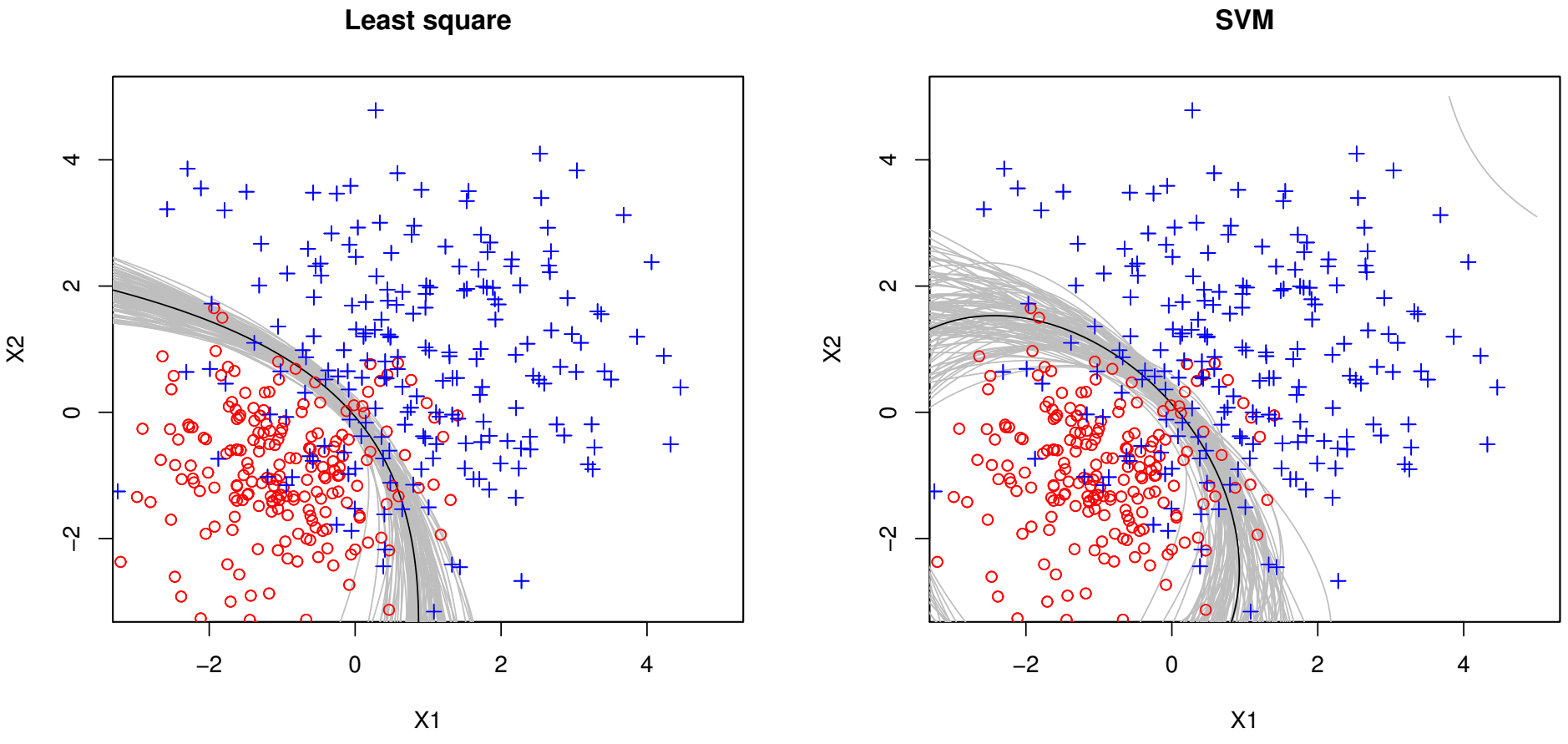

Here we illustrate the estimation procedure via a bivariate normal example with sample size $n = 400$. Assume the underlying distributions of the two classes are $f_1 = N((-1,-1)^T, I_2)$ and $f_2 = N((1,1)^T, 2I_2)$ with equal prior probability. We map the input $\{x_1, x_2\}$ to the polynomial basis $\{x_1, x_2, x_1x_2, x_1^2, x_2^2\}$ and fit the linear large-margin classifiers using the expanded inputs. The instability of the original nonlinear decision boundary boils down to the instability of the linear boundaries in the expanded space. Figure A1 demonstrates 100 nonlinear perturbed decision boundaries for the least squares and SVM losses, where the former is visually more stable than the latter. Indeed, their corresponding DBI estimations in the expanded space capture this relationship in that the estimated DBI of the former is 0.017 and that of the latter is 0.354. ∎

# Stability Enhanced Large-Margin Classifier Selection
## Supplementary Materials

Will Wei Sun[1], Guang Cheng[2], Yufeng Liu[3]

In this online supplementary note, we provide technical proofs of all lemmas, corollaries and theorems, discuss the calculation of the transformation matrix, and provide a notation table.

## S.1. Proof of Theorem 1:

Before we prove Theorem 1, we show an intermediate result in Lemma 2.

**Lemma 2** *Suppose Assumptions (L1)–(L3) hold and* $\lambda_n = o(n^{-1/2})$. *Then we have* $\widehat{\boldsymbol{\theta}}_L \xrightarrow{P} \boldsymbol{\theta}_{0L}$ *and* $\widehat{D}_L \xrightarrow{P} D_{0L}$ *as* $n \to \infty$.

To show $\widehat{\boldsymbol{\theta}}_L \to \boldsymbol{\theta}_{0L}$, we apply Theorem 5.7 of van der Vaart (1998). Firstly, we show that, uniformly in $\boldsymbol{\theta}$, the empirical risk $O_{nL}(\boldsymbol{\theta})$ converges to the true risk $\mathcal{R}_L(\boldsymbol{\theta})$ in probability. Assumption (L3) guarantees that the loss function $L(yf(\boldsymbol{x};\boldsymbol{\theta}))$ is convex in $\boldsymbol{\theta}$, and it is easy to see that $O_{nL}(\boldsymbol{\theta})$ converges to $\mathcal{R}_L(\boldsymbol{\theta})$ for each $\boldsymbol{\theta}$. Then we have $\sup_{\boldsymbol{\theta}} |O_{nL}(\boldsymbol{\theta}) - \mathcal{R}_L(\boldsymbol{\theta})| \to 0$ in probability by uniform convergence Theorem for convex functions in Pollard (1991). Secondly, according to assumption (L2), we have that $\mathcal{R}_L(\boldsymbol{\theta})$ has a unique minimizer $\boldsymbol{\theta}_{0L}$. Therefore, we know that $\widehat{\boldsymbol{\theta}}_L$ converges to $\boldsymbol{\theta}_{0L}$ in probability. The consistency of $\widehat{D}(\widehat{\boldsymbol{\theta}}_L)$ can be obtained by the uniform law of large numbers. According to Assumption (L1), $p(\boldsymbol{x})$ is continuously differentiable, and hence $|y - \text{sign}\{f(\boldsymbol{x};\boldsymbol{\theta})\}| = |y - \text{sign}\{\tilde{\boldsymbol{x}}^T\boldsymbol{\theta}\}|$ is continuous in each $\boldsymbol{\theta}$ for almost all $\boldsymbol{x}$. This together with $|y - \text{sign}\{f(\boldsymbol{x};\boldsymbol{\theta})\}| \le 2$ leads to uniform convergence $\sup_{\boldsymbol{\theta}} |\widehat{D}(\boldsymbol{\theta}) - \frac{1}{2}E|y_0 - \text{sign}\{f(\boldsymbol{x}_0;\boldsymbol{\theta})\}|| \to 0$. Therefore, we have $\widehat{D}(\widehat{\boldsymbol{\theta}}_L) \to D_{0L}$ in probability. This concludes the proof of Lemma 2. ■

**Proof of Theorem 1**:

We next prove (4) in three steps. Let $M_i(\boldsymbol{\theta}_{0L}) = \nabla_{\boldsymbol{\theta}} L(Y_i f(\boldsymbol{X}_i;\boldsymbol{\theta}))|_{\boldsymbol{\theta}=\boldsymbol{\theta}_{0L}}$. In step 1, we show that

$$\sqrt{n}(\widehat{\boldsymbol{\theta}}_L - \boldsymbol{\theta}_{0L}) = -n^{-1/2} H(\boldsymbol{\theta}_{0L})^{-1} \sum_{i=1}^n M_i(\boldsymbol{\theta}_{0L}) + o_P(1) \tag{S.1}$$

by applying Theorem 2.1 in Hjort and Pollard (1993). Denote $Z = (\boldsymbol{X}^T, Y)$ and $\Delta\boldsymbol{\theta} = (\Delta b, \Delta\boldsymbol{w}^T)^T$. Taylor expansion leads to

$$L(Yf(\boldsymbol{X};\boldsymbol{\theta}_{0L} + \Delta\boldsymbol{\theta})) - L(Yf(\boldsymbol{X};\boldsymbol{\theta}_{0L})) = M(\boldsymbol{\theta}_{0L})^T\Delta\boldsymbol{\theta} + R(Z, \Delta\boldsymbol{\theta}), \tag{S.2}$$

[1]Assistant Professor, Department of Management Science, University of Miami, FL 33156, Email: wsun@bus.miami.edu. This work was carried out during Will's PhD period at Purdue University.

[2]Corresponding Author. Professor, Department of Statistics, Purdue University, West Lafayette, IN 47906, Email: chengg@purdue.edu. Partially supported by NSF Grant DMS-0906497, CAREER Award DMS-1151692, DMS-1418042 and Simons Foundation 305266. I acknowledge SAMSI and Princeton University for their hospitality; part of the work was done during my visits.

[3]Professor, Department of Statistics and Operations Research, Department of Genetics, Department of Biostatistics, Carolina Center for Genome Sciences, Lineberger Comprehensive Cancer Center, University of North Carolina Chapel Hill, NC 27599, Email: yfliu@email.unc.edu. Partially supported by NSF DMS-1407241, and NIH/NCI P01 CA-142538.

where

$$M(\boldsymbol{\theta}_{0L}) = \nabla_{\boldsymbol{\theta}} L(Yf(\boldsymbol{X};\boldsymbol{\theta}))\Big|_{\boldsymbol{\theta}=\boldsymbol{\theta}_{0L}}; \quad R(Z,\Delta\boldsymbol{\theta}) = \frac{(\Delta\boldsymbol{\theta})^T\left(\nabla^2_{\boldsymbol{\theta}} L(Yf(\boldsymbol{X};\boldsymbol{\theta}))\Big|_{\boldsymbol{\theta}=\boldsymbol{\theta}_{0L}}\right)\Delta\boldsymbol{\theta}}{2} + o(\|\Delta\boldsymbol{\theta}\|^2).$$

According to Assumption (L1), it is easy to check that $E(M(\boldsymbol{\theta}_{0L})) = \nabla_{\boldsymbol{\theta}}\mathcal{R}_L(\boldsymbol{\theta})|_{\boldsymbol{\theta}=\boldsymbol{\theta}_{0L}} = 0$, and

$$E[R(Z,\Delta\boldsymbol{\theta})] = \frac{1}{2}(\Delta\boldsymbol{\theta})^T H(\boldsymbol{\theta}_{0L})(\Delta\boldsymbol{\theta}) + o(\|\Delta\boldsymbol{\theta}\|^2); \quad E[R^2(Z,\Delta\boldsymbol{\theta})] = o(\|\Delta\boldsymbol{\theta}\|^3).$$

Denote $s = (b_s, \boldsymbol{w}_s^T)^T$, $Z_i = (\boldsymbol{X}_i^T, Y_i)$, and

$$\begin{aligned} A_n(s) &= \sum_{i=1}^n \Big\{L(Y_i f(\boldsymbol{X}_i;\boldsymbol{\theta}_{0L} + s/\sqrt{n})) - L(Y_i f(\boldsymbol{X}_i;\boldsymbol{\theta}_{0L}))\Big\} \\ &\quad + \lambda_n(\boldsymbol{w}_{0L} + \boldsymbol{w}_s/\sqrt{n})^T(\boldsymbol{w}_{0L} + \boldsymbol{w}_s/\sqrt{n}) - \lambda_n \boldsymbol{w}_{0L}^T\boldsymbol{w}_{0L}. \end{aligned}$$

Note that $A_n(s)$ is minimized when $s = \sqrt{n}(\widehat{\boldsymbol{\theta}}_L - \boldsymbol{\theta}_{0L})$ and $nE[R(Z,s/\sqrt{n})] = \frac{1}{2}s^T H(\boldsymbol{\theta}_{0L})s + o(\|s\|^2)$. Based on the above Taylor expansion (S.2), we have

$$\begin{aligned} A_n(s) &= \sum_{i=1}^n \Big\{M_i(\boldsymbol{\theta}_{0L})^T s/\sqrt{n} + R(Z_i, s/\sqrt{n}) - ER(Z_i, s/\sqrt{n})\Big\} + nE[R(Z,s/\sqrt{n})] + \lambda_n \boldsymbol{w}_s^T\boldsymbol{w}_s \\ &= U_n^T s + \frac{1}{2}s^T H(\boldsymbol{\theta}_{0L})s + o(\|s\|^2) + \sum_{i=1}^n \Big\{R(Z_i, s/\sqrt{n}) - ER(Z_i, s/\sqrt{n})\Big\} + \lambda_n \boldsymbol{w}_s^T\boldsymbol{w}_s, \end{aligned}$$

where $U_n = n^{-1/2}\sum_{i=1}^n M_i(\boldsymbol{\theta}_{0L})$. Note that $\sum_{i=1}^n\{R(Z_i, s/\sqrt{n}) - ER(Z_i, s/\sqrt{n})\} \to 0$, and $\lambda_n \boldsymbol{w}_s^T\boldsymbol{w}_s \to 0$ since $\lambda_n \to 0$ and $\boldsymbol{w}_s$ is bounded. In addition, Hessian matrix $H(\boldsymbol{\theta}_{0L})$ is positive definite due to Assumption (L5). Therefore, we can conclude that (S.1) holds by Theorem 2.1 in Hjort and Pollard (1993).

In step 2, we show that $W_L = \sqrt{n}\{\widehat{D}(\widehat{\boldsymbol{\theta}}_L) - D_{0L}\} \to N(0, E(\psi_1^2))$. As shown in Jiang et al. (2008), the class of functions $\mathcal{G}_{\boldsymbol{\theta}}(\delta) = \Big\{|Y - \text{sign}\{f(\boldsymbol{X};\boldsymbol{\theta})\}| : \|\boldsymbol{\theta} - \boldsymbol{\theta}_{0L}\| \le \delta\Big\}$ is a P-Donsker class for any fixed $0 < \delta < \infty$. This together with (*S*.1) and consistency of $\widehat{\boldsymbol{\theta}}_L$ implies that

$$\begin{aligned} & \sqrt{n}\Big(\widehat{D}(\widehat{\boldsymbol{\theta}}_L) - D_{0L}\Big) \\ =\ & \sqrt{n}\Big(\widehat{D}(\widehat{\boldsymbol{\theta}}_L) - \widehat{D}(\boldsymbol{\theta}_{0L})\Big) + \sqrt{n}\Big(\widehat{D}(\boldsymbol{\theta}_{0L}) - D_{0L}\Big) \\ \stackrel{d}{=}\ & \sqrt{n}\dot{d}(\boldsymbol{\theta}_{0L})^T(\widehat{\boldsymbol{\theta}}_L - \boldsymbol{\theta}_{0L}) + \sqrt{n}\Big(\widehat{D}(\boldsymbol{\theta}_{0L}) - D_{0L}\Big) \\ \stackrel{d}{=}\ & n^{-1/2}\sum_{i=1}^n \Big\{\frac{1}{2}|Y_i - \text{sign}\{f(\boldsymbol{X}_i;\boldsymbol{\theta}_{0L})\}| - D_{0L} - \dot{d}(\boldsymbol{\theta}_{0L})^T H(\boldsymbol{\theta}_{0L})^{-1} M_1(\boldsymbol{\theta}_{0L})\Big\} \\ =\ & n^{-1/2}\sum_{i=1}^n \psi_i \stackrel{d}{\longrightarrow} N(0, E(\psi_1^2)), \end{aligned}$$

where "$\stackrel{d}{=}$" means asymptotical equivalence in the distributional sense.

In step 3, the distribution of $\mathcal{W}_L = n^{1/2}\{\widehat{\mathcal{D}}_L - D_{0L}\}$ is asymptotically equivalent to that of $W_L$ as shown in Theorem 3 in Jiang et al. (2008). This concludes the proof of Theorem 1. ■

## S.2. Proof of Theorem 2

According to Appendix D in Jiang et al. (2008), we have

$$W_1^* \stackrel{d}{=} n^{-1/2}\sum_{i=1}^n \psi_{i1}(G_i - 1) \text{ and } W_2^* \stackrel{d}{=} n^{-1/2}\sum_{i=1}^n \psi_{i2}(G_i - 1),$$

where $\psi_{ij} = \frac{1}{2}|Y_i - \text{sign}\{f(\boldsymbol{X}_i;\boldsymbol{\theta}_{0j})\}| - D_{0j} - \dot{d}(\boldsymbol{\theta}_{0j})^T H(\boldsymbol{\theta}_{0j})^{-1} M_i(\boldsymbol{\theta}_{0j})$, for $j = 1, 2$. Recall that "$\stackrel{d}{=}$" means the distributional equivalence. As shown in Jiang et al. (2008), conditional on the data, $W_j^*$ converges to a normal with mean 0 and variance $n^{-1}\sum_{i=1}^n \psi_{ij}^2$ for $j = 1, 2$. Note that

$$W_2^* - W_1^* \stackrel{d}{=} n^{-1/2}\sum_{i=1}^n (\psi_{i2} - \psi_{i1})(G_i - 1).$$

Here, $(\psi_{i2}-\psi_{i1})$'s, $i = 1, \ldots, n$, are i.i.d random vectors with $E(\psi_{i2}-\psi_{i1}) = 0$ and $E|\psi_{i2}-\psi_{i1}|^2 < \infty$. Independent of $(\psi_{i2} - \psi_{i1})$, $(G_i - 1)$'s are i.i.d random variables with mean 0 and variance 1. Since $(\psi_{i2}-\psi_{i1})$ depends on the sample $(\boldsymbol{x}_i, y_i)$, Lemma 2.9.5 in van der Vaart and Wellner (1996) implies that, conditional on the data,

$$n^{-1/2}\sum_{i=1}^n (\psi_{i2} - \psi_{i1})(G_i - 1) \stackrel{d}{\Longrightarrow} N(0, Var(\psi_{12} - \psi_{11})). \tag{S.3}$$

Next, as shown in Theorem 1, $W_1 \stackrel{d}{=} n^{-1/2}\sum_{i=1}^n \psi_{i1}$ and $W_2 \stackrel{d}{=} n^{-1/2}\sum_{i=1}^n \psi_{i2}$, therefore,

$$W_2 - W_1 \stackrel{d}{=} n^{-1/2}\sum_{i=1}^n (\psi_{i2} - \psi_{i1}) \stackrel{d}{\longrightarrow} N(0, Var(\psi_{12} - \psi_{11})).$$

This together with (S.3) and the asymptotic equivalence of $W_L$ and $\mathcal{W}_L$ (Jiang et al. 2008) lead to the asymptotic equivalence between $\mathcal{W}_{\Delta 12}$ and $W^*_{\Delta_{12}}$, which concludes the proof. ■

## S.3. Calculation of the transformation matrix in Section 3.2

Given a $d$ dimensional hyperplane $f(\boldsymbol{x};\boldsymbol{\theta}) = b + w_1 x_1 + \cdots + w_d x_d = 0$, we aim to find a transformation matrix $R \in \mathbb{R}^{d\times d}$ such that the transformed hyperplane $f(\boldsymbol{x};\boldsymbol{\theta}^\dagger) = b^\dagger + w_1^\dagger x_1 + \cdots + w_d^\dagger x_d = 0$ is parallel to $\mathcal{X}_1, \ldots, \mathcal{X}_{d-1}$, where $(w_1^\dagger, \cdots, w_d^\dagger)^T = R(w_1, \cdots, w_d)^T$ and $b^\dagger = b$. Here, we implicitly assume that $w_d \neq 0$.

We construct a class of linearly independent vectors spanning the hyperplane:

$$\begin{bmatrix} 1 \\ 0 \\ \vdots \\ 0 \\ -\frac{w_1}{w_d} \end{bmatrix} \begin{bmatrix} 0 \\ 1 \\ \vdots \\ 0 \\ -\frac{w_2}{w_d} \end{bmatrix} \cdots \begin{bmatrix} 0 \\ 0 \\ \vdots \\ 1 \\ -\frac{w_{d-1}}{w_d} \end{bmatrix}.$$

Denote these vectors as $v_1$, $v_2$,...,$v_{d-1}$. Then, by Gram-Schmidt process, we can produce the following orthogonal vectors $\bar{v}_1$, $\bar{v}_2$,..., $\bar{v}_{d-1}$:

$$\begin{aligned} \bar{v}_1 &= v_1, \\ \bar{v}_2 &= v_2 - \tfrac{<v_2,\bar{v}_1>}{<\bar{v}_1,\bar{v}_1>}\bar{v}_1, \\ \bar{v}_{d-1} &= v_{d-1} - \tfrac{<v_{d-1},\bar{v}_1>}{<\bar{v}_1,\bar{v}_1>}\bar{v}_1 - \cdots - \tfrac{<v_{d-1},\bar{v}_{d-2}>}{<\bar{v}_{d-2},\bar{v}_{d-2}>}\bar{v}_{d-2}, \end{aligned}$$

where the inner product $< u, v >= \sum_{i=1}^{d} u_i v_i$ for $u = (u_1, \ldots, u_d)$ and $v = (v_1, \ldots, v_d)$. Denote $\bar{v}_d = [w_1, \cdots, w_d]^T$, which is orthogonal to every $\bar{v}_i$, $i = 1, \cdots, d-1$ by the above construction. In the end, we normalize $u_i = \bar{v}_i \|\bar{v}_i\|^{-1}$ for $i = 1, \cdots, d$, and define the orthogonal transformation matrix $R$ as $[u_1, \ldots, u_d]^T$. By some elementary calculation, we can verify that that $w_i^\dagger = 0$ for $i = 1, \cdots, d-1$ but $w_d^\dagger \neq 0$ under the above construction. Therefore, the transformed hyperplane $f(\boldsymbol{x}; \boldsymbol{\theta}^\dagger)$ is parallel to $\mathcal{X}_1, \ldots, \mathcal{X}_{d-1}$. ■

## S.4. Asymptotic Normality of $\widehat{\boldsymbol{\theta}}_\gamma$ and $\widehat{\mathcal{D}}_\gamma$ for LUM

This section establishes the asymptotic normality of $\widehat{\mathcal{D}}_\gamma$ and $\widehat{\boldsymbol{\theta}}_\gamma$ (with more explicit forms of the asymptotic variances) by verifying the conditions in Theorem 1, i.e., (L1)–(L5). In particular, we provide a set of sufficient conditions for the LUM, i.e., (L1) and (A1) below.

(A1) $\text{Var}(\boldsymbol{X}|Y) \in \mathbb{R}^{d\times d}$ is a positive definite matrix for $Y \in \{1, -1\}$.

Assumption (A1) is needed to guarantee the uniqueness of the true minimizer $\boldsymbol{\theta}_{0\gamma}$. It is worth pointing out that the asymptotic normality of the estimated coefficients for SVM has also been established by Koo et al. (2008) under another set of assumptions.

**Corollary 1** *Suppose Assumptions (L1) and (A1) hold and $\lambda_n = o(n^{-1/2})$. For each $\gamma \in [0, 1]$,*

$$\sqrt{n}(\widehat{\boldsymbol{\theta}}_\gamma - \boldsymbol{\theta}_{0\gamma}) \xrightarrow{d} N(0, \Sigma_{0\gamma}) \quad \text{as } n \to \infty, \tag{S.4}$$

*where $\Sigma_{0\gamma} = H(\boldsymbol{\theta}_{0\gamma})^{-1} G(\boldsymbol{\theta}_{0\gamma}) H(\boldsymbol{\theta}_{0\gamma})^{-1}$ with $G(\boldsymbol{\theta}_{0\gamma})$ and $H(\boldsymbol{\theta}_{0\gamma})$ defined in (S.6) and (S.10) in the supplementary materials.*

In practice, direct estimation of $\Sigma_{0\gamma}$ in (*S.4*) is difficult because of the involvement of the Dirac delta function; see (*S.8*) and (*S.9*) in Section S.5 of the supplementary materials. Instead, we find that the perturbation-based resampling procedure proposed in Stage 1 works well.

Next we establish the asymptotic normality of $\widehat{\mathcal{D}}_\gamma$.

**Corollary 2** *Suppose that the assumptions in Corollary 1 hold. We have, as $n \to \infty$,*

$$\sqrt{n}(\widehat{\mathcal{D}}_\gamma - D_{0\gamma}) \xrightarrow{d} N\Big(0, E(\psi_{1\gamma}^2)\Big), \tag{S.5}$$

*where* $\psi_{1\gamma} = \frac{1}{2}|Y_1 - sign\{f(\boldsymbol{X}_1;\boldsymbol{\theta}_{0\gamma})\}| - D_{0\gamma} - \dot{d}(\boldsymbol{\theta}_{0\gamma})^T H(\boldsymbol{\theta}_{0\gamma})^{-1} M_1(\boldsymbol{\theta}_{0\gamma})$, $\dot{d}(\boldsymbol{\theta}) = \triangledown_{\boldsymbol{\theta}} E(\widehat{D}_\gamma(\boldsymbol{\theta}))$, *and*

$$M_1(\boldsymbol{\theta}_{0\gamma}) = -Y_1\tilde{\boldsymbol{X}}_1 I_{\{Y_1 f(\boldsymbol{X}_1;\boldsymbol{\theta}_{0\gamma})<\gamma\}} - \frac{(1-\gamma)^2 Y_1 \tilde{\boldsymbol{X}}_1 I_{\{Y_1 f(\boldsymbol{X}_1;\boldsymbol{\theta}_{0\gamma})\geq\gamma\}}}{\Big(Y_1 f(\boldsymbol{X}_1;\boldsymbol{\theta}_{0\gamma}) - 2\gamma + 1\Big)^2}.$$

Corollary 2 demonstrates that the K-CV error induced from each LUM loss function yields the desirable asymptotic property under Assumptions (L1) and (A1). It can be applied to justify the perturbation-based resampling procedure for LUM as shown in Theorem 2.

## S.5. Proof of Corollary 1

It suffices to show that (A1) and (L1) imply Assumptions (L2)-(L5).

(L2). We first show that the minimizer $\boldsymbol{\theta}_{0\gamma}$ exists for each fixed $\gamma$. It is easy to see that $\mathcal{R}_\gamma(\boldsymbol{\theta})$ is continuous w.r.t. $\boldsymbol{\theta}$. We next show that, for any large enough $M$, the closed set $S(M) = \Big\{\boldsymbol{\theta} \in R^d : \mathcal{R}_\gamma(\boldsymbol{\theta}) \leq M\Big\}$ is bounded. When $yf(\boldsymbol{x},\boldsymbol{\theta}) < \gamma$, we need to show $S(M) = \Big\{\boldsymbol{\theta} \in R^d : E[1 - Yf(\boldsymbol{X};\boldsymbol{\theta})] \leq M\Big\}$ is contained in a box around the origin. Denote $e_j$ as the vector with one in the $j$-th component and zero otherwise. Motivated by Rocha et al. (2009), we can show that, for any $M$, there exists a $\alpha_{j,M}$ such that any $\boldsymbol{\theta}$ satisfying $| < \boldsymbol{\theta}, e_j > | > \alpha_{j,M}$ leads to $E[(1 - Yf(\boldsymbol{X};\boldsymbol{\theta})I_{(Yf(\boldsymbol{X};\boldsymbol{\theta})<\gamma)})] > M$. Similarly, when $yf(\boldsymbol{x},\boldsymbol{\theta}) \geq \gamma$, $S(M)$ is contained in a sphere around the origin, that is, for any $M$, there exists a $\sigma$ such that any $\boldsymbol{\theta}$ satisfying $| < \boldsymbol{\theta}, \boldsymbol{\theta} > | > \sigma$ leads to $E[\frac{(1-\gamma)^2}{Yf(\boldsymbol{X};\boldsymbol{\theta})-2\gamma+1} I_{(Yf(\boldsymbol{X};\boldsymbol{\theta})\geq\gamma)})] > M$. These imply the existence of $\boldsymbol{\theta}_{0\gamma}$. The uniqueness of $\boldsymbol{\theta}_{0\gamma}$ is implied by the positive definiteness of Hessian matrix as verified in (L5) below.

(L3). The loss function $L_\gamma(yf(\boldsymbol{x};\boldsymbol{\theta}))$ is convex by noting that two segments of $L_\gamma(yf(\boldsymbol{x};\boldsymbol{\theta}))$ are convex, and the sum of convex functions is convex.

(L4). The loss function $L_\gamma(yf(\boldsymbol{x};\boldsymbol{\theta}))$ is not differentiable only on the set $\{\boldsymbol{x} : \tilde{\boldsymbol{x}}^T\boldsymbol{\theta} = \gamma \text{ or } \tilde{\boldsymbol{x}}^T\boldsymbol{\theta} = -\gamma\}$, which is assumed to be a zero probability event. Therefore, with probability one, it is differentiable with

$$\triangledown_{\boldsymbol{\theta}} L_\gamma(yf(\boldsymbol{x};\boldsymbol{\theta})) = -\tilde{\boldsymbol{x}} y I_{(y\tilde{\boldsymbol{x}}^T\boldsymbol{\theta}<\gamma)} - \frac{(1-\gamma)^2\tilde{\boldsymbol{x}}y}{(y\tilde{\boldsymbol{x}}^T\boldsymbol{\theta} - 2\gamma+1)^2} I_{(y\tilde{\boldsymbol{x}}^T\boldsymbol{\theta}\geq\gamma)},$$

and hence

$$\begin{aligned} G(\boldsymbol{\theta}_{0\gamma}) &= E\Big[\triangledown_{\boldsymbol{\theta}} L_\gamma(Yf(\boldsymbol{X};\boldsymbol{\theta}))\triangledown_{\boldsymbol{\theta}} L_\gamma(Yf(\boldsymbol{X};\boldsymbol{\theta}))^T|_{\boldsymbol{\theta}=\boldsymbol{\theta}_{0\gamma}}\Big] \\ &= E\Big\{\tilde{\boldsymbol{X}}\tilde{\boldsymbol{X}}^T Y^2 I_{(Y\tilde{\boldsymbol{X}}^T\boldsymbol{\theta}_{0\gamma}<\gamma)} + \frac{(1-\gamma)^4\tilde{\boldsymbol{X}}\tilde{\boldsymbol{X}}^T Y^2}{(Y\tilde{\boldsymbol{X}}^T\boldsymbol{\theta}_{0\gamma} - 2\gamma+1)^4} I_{(Y\tilde{\boldsymbol{X}}^T\boldsymbol{\theta}_{0\gamma}\geq\gamma)}\Big\} \\ &= E\Big\{\tilde{\boldsymbol{X}}\tilde{\boldsymbol{X}}^T\Big[p(\boldsymbol{X})A(\boldsymbol{X},\boldsymbol{\theta}_{0\gamma}) + (1-p(\boldsymbol{X}))B(\boldsymbol{X},\boldsymbol{\theta}_{0\gamma})\Big]\Big\}, \end{aligned} \tag{S.6}$$

where $A(\boldsymbol{X}, \boldsymbol{\theta}_{0\gamma})$ and $B(\boldsymbol{X}, \boldsymbol{\theta}_{0\gamma})$ are defined as

$$\begin{aligned} A(\boldsymbol{X}, \boldsymbol{\theta}_{0\gamma}) &= I_{(\tilde{\boldsymbol{X}}^T\boldsymbol{\theta}_{0\gamma}<\gamma)} + \frac{(1-\gamma)^4}{(\tilde{\boldsymbol{X}}^T\boldsymbol{\theta}_{0\gamma}-2\gamma+1)^4} I_{(\tilde{\boldsymbol{X}}^T\boldsymbol{\theta}_{0\gamma}\geq\gamma)}; \\ B(\boldsymbol{X}, \boldsymbol{\theta}_{0\gamma}) &= I_{(-\tilde{\boldsymbol{X}}^T\boldsymbol{\theta}_{0\gamma}<\gamma)} + \frac{(1-\gamma)^4}{(\tilde{\boldsymbol{X}}^T\boldsymbol{\theta}_{0\gamma}+2\gamma-1)^4} I_{(-\tilde{\boldsymbol{X}}^T\boldsymbol{\theta}_{0\gamma}\geq\gamma)}. \end{aligned}$$

Obviously, $|A(\boldsymbol{X}, \boldsymbol{\theta}_{0\gamma})|$ and $|B(\boldsymbol{X}, \boldsymbol{\theta}_{0\gamma})|$ are both bounded by one. Therefore, $G(\boldsymbol{\theta}_{0\gamma}) < \infty$ based on the moment condition of $\boldsymbol{X}$.

(L5). We prove it in three steps. First, we show the risk $R_\gamma(\boldsymbol{\theta})$ is bounded. For each fixed $\gamma \in [0, 1]$,

$$\begin{aligned} \mathcal{R}_\gamma(\boldsymbol{\theta}) \leq E\Big|L_\gamma(Yf(\boldsymbol{X};\boldsymbol{\theta}))\Big| &= E\Big|(1-Y\tilde{\boldsymbol{X}}^T\boldsymbol{\theta})I_{(Y\tilde{\boldsymbol{X}}^T\boldsymbol{\theta}<\gamma)} + \frac{(1-\gamma)^2}{Y\tilde{\boldsymbol{X}}^T\boldsymbol{\theta}-2\gamma+1} I_{(Y\tilde{\boldsymbol{X}}^T\boldsymbol{\theta}\geq\gamma)}\Big| \\ &\leq E\Big|(1-Y\tilde{\boldsymbol{X}}^T\boldsymbol{\theta})I_{(Y\tilde{\boldsymbol{X}}^T\boldsymbol{\theta}<\gamma)}\Big| + E\Big|\frac{(1-\gamma)^2}{Y\tilde{\boldsymbol{X}}^T\boldsymbol{\theta}-2\gamma+1} I_{(Y\tilde{\boldsymbol{X}}^T\boldsymbol{\theta}\geq\gamma)}\Big| \\ &\leq E\Big|(1-Y\tilde{\boldsymbol{X}}^T\boldsymbol{\theta})I_{(Y\tilde{\boldsymbol{X}}^T\boldsymbol{\theta}<1)}\Big| + |1-\gamma| < \infty, \end{aligned} \tag{S.7}$$

where the first term in (*S.7*) was shown to be bounded in Rocha et al. (2009).

Next, we derive the form of Hessian matrix. The moment assumption of $\boldsymbol{x}$ and the inequality $(y\tilde{\boldsymbol{x}}^T\boldsymbol{\theta} - 2\gamma + 1)^2 \leq (1-\gamma)^2$ lead to $E|\nabla_{\boldsymbol{\theta}} L_\gamma(Yf(\boldsymbol{X};\boldsymbol{\theta}))| \leq E|-\tilde{\boldsymbol{X}}Y| + E|-\tilde{\boldsymbol{X}}Y| \leq 2E|\tilde{\boldsymbol{X}}| < \infty$. Then, dominated convergence theorem implies that $\nabla_{\boldsymbol{\theta}}\mathcal{R}_\gamma(\boldsymbol{\theta}) = E[\nabla_{\boldsymbol{\theta}} L_\gamma(Yf(\boldsymbol{X};\boldsymbol{\theta}))]$. Hence, the Hessian matrix equals $\nabla_{\boldsymbol{\theta}} E[\nabla_{\boldsymbol{\theta}} L_\gamma(Yf(\boldsymbol{X};\boldsymbol{\theta}))]$. We next derive the form of $E[\nabla_{\boldsymbol{\theta}} L_\gamma(Yf(\boldsymbol{X};\boldsymbol{\theta}))]$. Note that

$$\begin{aligned} E[\nabla_{\boldsymbol{\theta}} L_\gamma(Yf(\boldsymbol{X};\boldsymbol{\theta}))] &= E\Big[-\tilde{\boldsymbol{X}}YI_{\{Y\tilde{\boldsymbol{X}}^T\boldsymbol{\theta}<\gamma\}} - \frac{(1-\gamma)^2\tilde{\boldsymbol{X}}Y}{(Y\tilde{\boldsymbol{X}}^T\boldsymbol{\theta}-2\gamma+1)^2} I_{\{Y\tilde{\boldsymbol{X}}^T\boldsymbol{\theta}\geq\gamma\}}\Big] \\ &= E\Big\{I_{\{Y=1\}}\Big[-\tilde{\boldsymbol{X}}I_{\{\tilde{\boldsymbol{X}}^T\boldsymbol{\theta}<\gamma\}} - \frac{(1-\gamma)^2\tilde{\boldsymbol{X}}}{(\tilde{\boldsymbol{X}}^T\boldsymbol{\theta}-2\gamma+1)^2} I_{\{\tilde{\boldsymbol{X}}^T\boldsymbol{\theta}\geq\gamma\}}\Big] \\ &\qquad + I_{\{Y=-1\}}\Big[\tilde{\boldsymbol{X}}I_{\{-\tilde{\boldsymbol{X}}^T\boldsymbol{\theta}<\gamma\}} + \frac{(1-\gamma)^2\tilde{\boldsymbol{X}}}{(\tilde{\boldsymbol{X}}^T\boldsymbol{\theta}+2\gamma-1)^2} I_{\{-\tilde{\boldsymbol{X}}^T\boldsymbol{\theta}\geq\gamma\}}\Big]\Big\} \\ &= E\Big\{p(\boldsymbol{X})\Big[-\tilde{\boldsymbol{X}}I_{\{\tilde{\boldsymbol{X}}^T\boldsymbol{\theta}<\gamma\}} - \frac{(1-\gamma)^2\tilde{\boldsymbol{X}}}{(\tilde{\boldsymbol{X}}^T\boldsymbol{\theta}-2\gamma+1)^2} I_{\{\tilde{\boldsymbol{X}}^T\boldsymbol{\theta}\geq\gamma\}}\Big]\Big\} \\ &\qquad + E\Big\{(1-p(\boldsymbol{X}))\Big[\tilde{\boldsymbol{X}}I_{\{-\tilde{\boldsymbol{X}}^T\boldsymbol{\theta}<\gamma\}} + \frac{(1-\gamma)^2\tilde{\boldsymbol{X}}}{(\tilde{\boldsymbol{X}}^T\boldsymbol{\theta}+2\gamma-1)^2} I_{\{-\tilde{\boldsymbol{X}}^T\boldsymbol{\theta}\geq\gamma\}}\Big]\Big\} \\ &= E_1(\boldsymbol{\theta}) + E_2(\boldsymbol{\theta}). \end{aligned}$$

After tedious algebra, we can show

$$\begin{aligned} \nabla_{\boldsymbol{\theta}} E_1(\boldsymbol{\theta})|_{\boldsymbol{\theta}=\boldsymbol{\theta}_{0\gamma}} &= E\Big\{\tilde{\boldsymbol{X}}\tilde{\boldsymbol{X}}^T p(\boldsymbol{X})C(\boldsymbol{X}, \boldsymbol{\theta}_{0\gamma})\Big\}, \\ \nabla_{\boldsymbol{\theta}} E_2(\boldsymbol{\theta})|_{\boldsymbol{\theta}=\boldsymbol{\theta}_{0\gamma}} &= E\Big\{\tilde{\boldsymbol{X}}\tilde{\boldsymbol{X}}^T (1-p(\boldsymbol{X}))D(\boldsymbol{X}, \boldsymbol{\theta}_{0\gamma})\Big\}, \end{aligned}$$

where

$$C(\boldsymbol{X}, \boldsymbol{\theta}_{0\gamma}) = \delta(\gamma - \tilde{\boldsymbol{X}}^T \boldsymbol{\theta}_{0\gamma}) - \frac{(1-\gamma)^2 \delta(\tilde{\boldsymbol{X}}^T \boldsymbol{\theta}_{0\gamma} - \gamma)}{(\tilde{\boldsymbol{X}}^T \boldsymbol{\theta}_{0\gamma} - 2\gamma + 1)^2} + \frac{2(1-\gamma)^2 I_{(\tilde{\boldsymbol{X}}^T \boldsymbol{\theta}_{0\gamma} \geq \gamma)}}{(\tilde{\boldsymbol{X}}^T \boldsymbol{\theta}_{0\gamma} - 2\gamma + 1)^3}, \quad \text{(S.8)}$$

$$D(\boldsymbol{X}, \boldsymbol{\theta}_{0\gamma}) = \delta(\gamma + \tilde{\boldsymbol{X}}^T \boldsymbol{\theta}_{0\gamma}) - \frac{(1-\gamma)^2 \delta(\tilde{\boldsymbol{X}}^T \boldsymbol{\theta}_{0\gamma} + \gamma)}{(\tilde{\boldsymbol{X}}^T \boldsymbol{\theta}_{0\gamma} + 2\gamma - 1)^2} - \frac{2(1-\gamma)^2 I_{(-\tilde{\boldsymbol{X}}^T \boldsymbol{\theta}_{0\gamma} \geq \gamma)}}{(\tilde{\boldsymbol{X}}^T \boldsymbol{\theta}_{0\gamma} + 2\gamma - 1)^3}, \quad \text{(S.9)}$$

and $\delta(\cdot)$ is the Dirac delta function. Hence, we can write the Hessian matrix as

$$H(\boldsymbol{\theta}_{0\gamma}) = E\Big\{\tilde{\boldsymbol{X}}\tilde{\boldsymbol{X}}^T\Big[p(\boldsymbol{X})C(\boldsymbol{X}, \boldsymbol{\theta}_{0\gamma}) + (1 - p(\boldsymbol{X}))D(\boldsymbol{X}, \boldsymbol{\theta}_{0\gamma})\Big]\Big\}. \quad \text{(S.10)}$$

Finally, we establish the positive definiteness of $H(\boldsymbol{\theta}_{0\gamma})$. We write $H(\boldsymbol{\theta}_{0\gamma}) = R_1(\boldsymbol{\theta}_{0\gamma}) + R_2(\boldsymbol{\theta}_{0\gamma})$ with

$$R_1(\boldsymbol{\theta}_{0\gamma}) = E\Bigg\{\tilde{\boldsymbol{X}}\tilde{\boldsymbol{X}}^T\Big[p(\boldsymbol{X})\delta(\gamma - \tilde{\boldsymbol{X}}^T \boldsymbol{\theta}_{0\gamma}) + (1 - p(\boldsymbol{X}))\delta(\gamma + \tilde{\boldsymbol{X}}^T \boldsymbol{\theta}_{0\gamma})\Big]\Bigg\},$$

$$R_2(\boldsymbol{\theta}_{0\gamma}) = (1-\gamma)^2 E\Bigg\{\tilde{\boldsymbol{X}}\tilde{\boldsymbol{X}}^T\Big[(1 - p(\boldsymbol{X}))\Big(\frac{\delta(\gamma + \tilde{\boldsymbol{X}}^T \boldsymbol{\theta}_{0\gamma})}{(\tilde{\boldsymbol{X}}^T \boldsymbol{\theta}_{0\gamma} + 2\gamma - 1)^2} - \frac{2I_{(-\tilde{\boldsymbol{X}}^T \boldsymbol{\theta}_{0\gamma} \geq \gamma)}}{(\tilde{\boldsymbol{X}}^T \boldsymbol{\theta}_{0\gamma} + 2\gamma - 1)^3}\Big)$$
$$- p(\boldsymbol{X})\Big(\frac{\delta(\tilde{\boldsymbol{X}}^T \boldsymbol{\theta}_{0\gamma}) - \gamma}{(\tilde{\boldsymbol{X}}^T \boldsymbol{\theta}_{0\gamma} - 2\gamma + 1)^2} - \frac{2I_{(\tilde{\boldsymbol{X}}^T \boldsymbol{\theta}_{0\gamma} \leq \gamma)}}{(\tilde{\boldsymbol{X}}^T \boldsymbol{\theta}_{0\gamma} - 2\gamma + 1)^3}\Big)\Big]\Bigg\}.$$

Next we show the positive definiteness of $R_1(\boldsymbol{\theta}_{0\gamma})$. Let $f_{\boldsymbol{x}}$ be the density of $\tilde{\boldsymbol{x}}^T \boldsymbol{\theta}_{0\gamma}$. According to Lemma 9 in Rocha et al. (2009), Assumption $(L1)$ implies that $f_{\boldsymbol{x}}(\gamma) > 0$, $f_{\boldsymbol{x}}(-\gamma) > 0$, $P(Y = 1|\tilde{\boldsymbol{X}}^T \boldsymbol{\theta}_{0\gamma} = \gamma) > 0$, and $P(Y = -1|\tilde{\boldsymbol{X}}^T \boldsymbol{\theta}_{0\gamma} = -\gamma) > 0$. Note that $R_1(\boldsymbol{\theta}_{0\gamma})$ can be rewritten as

$$\begin{aligned} R_1(\boldsymbol{\theta}_{0\gamma}) &= E\Big[\tilde{\boldsymbol{X}}\tilde{\boldsymbol{X}}^T|Y = 1, \tilde{\boldsymbol{X}}^T \boldsymbol{\theta}_{0\gamma} = \gamma\Big]P(Y = 1|\tilde{\boldsymbol{X}}^T \boldsymbol{\theta}_{0\gamma} = \gamma) f_{\boldsymbol{X}}(\gamma) \\ &\quad + E\Big[\tilde{\boldsymbol{X}}\tilde{\boldsymbol{X}}^T|Y = -1, \tilde{\boldsymbol{X}}^T \boldsymbol{\theta}_{0\gamma} = -\gamma\Big]P(Y = -1|\tilde{\boldsymbol{X}}^T \boldsymbol{\theta}_{0\gamma} = -\gamma) f_{\boldsymbol{X}}(-\gamma). \end{aligned}$$

In order to show $R_1(\boldsymbol{\theta}_{0\gamma})$ is positive definite, it remains to show that $E\Big[\tilde{\boldsymbol{X}}\tilde{\boldsymbol{X}}^T|Y = 1, \tilde{\boldsymbol{X}}^T \boldsymbol{\theta}_{0\gamma} = \gamma\Big]$ or $E\Big[\tilde{\boldsymbol{X}}\tilde{\boldsymbol{X}}^T|Y = -1, \tilde{\boldsymbol{X}}^T \boldsymbol{\theta}_{0\gamma} = -\gamma\Big]$ is strictly positive definite. Rocha et al. (2009) showed that

$$\begin{aligned} E\Big[\tilde{\boldsymbol{X}}\tilde{\boldsymbol{X}}^T|Y, \tilde{\boldsymbol{X}}^T \theta_{0\gamma} = \gamma\Big] &= E\Big[\tilde{\boldsymbol{X}}\tilde{\boldsymbol{X}}^T|Y, \boldsymbol{X}^T v_{\boldsymbol{w}_{0\gamma}} = \frac{\gamma - b_{0\gamma}}{\|\boldsymbol{w}_{0\gamma}\|}\Big] \\ &\succeq \Big(\frac{\gamma - b_{0\gamma}}{\|\boldsymbol{w}_{0\gamma}\|}\Big)^2 (v_{\boldsymbol{w}_{0\gamma}} v_{\boldsymbol{w}_{0\gamma}}^T) + Var\Big(\boldsymbol{X}|Y, \boldsymbol{X}^T v_{\boldsymbol{w}_{0\gamma}} = \frac{\gamma - b_{0\gamma}}{\|\boldsymbol{w}_{0\gamma}\|}\Big), \end{aligned} \quad \text{(S.11)}$$

where $S_1 \succeq S_2$ means $S_1 - S_2$ is positive semi-definite, and $v_{\boldsymbol{w}_{0\gamma}} = \frac{\boldsymbol{w}_{0\gamma}}{\|\boldsymbol{w}_{0\gamma}\|}$. By assumption $(A1)$, $Var(\boldsymbol{X}|Y)$ is non-singular, and hence $Var\Big(\boldsymbol{X}|Y, \boldsymbol{X}^T v_{\boldsymbol{w}_{0\gamma}} = \frac{\gamma - b_{0\gamma}}{\|\boldsymbol{w}_{0\gamma}\|}\Big)$ has rank $(d-1)$ . Therefore, the right hand side of (S.11) is strictly positive definite when $\gamma \neq b_{0\gamma}$. Similarly, $E\Big[\tilde{\boldsymbol{X}}\tilde{\boldsymbol{X}}^T|Y, \tilde{\boldsymbol{X}}^T \theta_{0\gamma} =$ $-\gamma\Big]$ is strictly positive definite when $\gamma \neq -b_{0\gamma}$. Therefore, either $E\Big[\tilde{\boldsymbol{X}}\tilde{\boldsymbol{X}}^T|Y=1, \boldsymbol{X}^T\boldsymbol{w}_{0\gamma}+b_{0\gamma} = \gamma\Big]$ or $E\Big[\tilde{\boldsymbol{X}}\tilde{\boldsymbol{X}}^T|Y=-1, \boldsymbol{X}^T\boldsymbol{w}_{0\gamma}+b_{0\gamma} = -\gamma\Big]$ will be strictly positive definite at $\boldsymbol{\theta}_{0\gamma}$. This leads to the positive definiteness of $R_1(\boldsymbol{\theta}_{0\gamma})$.

In addition, similar argument implies that $R_2(\boldsymbol{\theta}_{0\gamma})$ is positive definite at $\boldsymbol{\theta}_{0\gamma}$. This is due to the fact that $(\tilde{\boldsymbol{X}}^T\boldsymbol{\theta}_{0\gamma}+2\gamma-1)^3 < 0$ when $\tilde{\boldsymbol{X}}^T\boldsymbol{\theta}_{0\gamma}+\gamma \leq 0$, and $(\tilde{\boldsymbol{X}}^T\boldsymbol{\theta}_{0\gamma}-2\gamma+1)^3 > 0$ when $\tilde{\boldsymbol{X}}^T\boldsymbol{\theta}_{0\gamma}-\gamma \geq 0$. Therefore, the Hessian matrix $H(\boldsymbol{\theta}_{0\gamma})$ is strictly positive definite for any $\gamma \in [0,1]$. This concludes the proof of Corollary 1. ■

## S.6. Proof of Corollary 2

Following the proof of Theorem 1, we only need to show that

$$\sqrt{n}(\widehat{\boldsymbol{\theta}}_\gamma - \boldsymbol{\theta}_{0\gamma}) = -n^{-1/2}H(\boldsymbol{\theta}_{0\gamma})^{-1}\sum_{i=1}^n M_i(\boldsymbol{\theta}_{0\gamma}) + o_P(1),$$

where

$$M_i(\boldsymbol{\theta}_{0\gamma}) = -Y_i\tilde{\boldsymbol{X}}_i I_{\{Y_i f(\boldsymbol{X}_i;\boldsymbol{\theta}_{0\gamma})<\gamma\}} - \frac{(1-\gamma)^2 Y_i\tilde{\boldsymbol{X}}_i I_{\{Y_i f(\boldsymbol{X}_i;\boldsymbol{\theta}_{0\gamma})\geq\gamma\}}}{\Big(Y_i f(\boldsymbol{X}_i;\boldsymbol{\theta}_{0\gamma})-2\gamma+1\Big)^2}.$$

Similarly, we denote $Z=(\boldsymbol{X}^T, Y)$ and $t=(b_t, \boldsymbol{w}_t^T)^T$, and write

$$\begin{aligned}
& L_\gamma(Yf(\boldsymbol{X};\boldsymbol{\theta}_{0\gamma}+t)) - L_\gamma(Yf(\boldsymbol{X};\boldsymbol{\theta}_{0\gamma})) \\
&= (1-Y\tilde{\boldsymbol{X}}^T(\boldsymbol{\theta}_{0\gamma}+t))I_{\{Y\tilde{\boldsymbol{X}}^T(\boldsymbol{\theta}_{0\gamma}+t)<\gamma\}} + \frac{(1-\gamma)^2}{Y\tilde{\boldsymbol{X}}^T(\boldsymbol{\theta}_{0\gamma}+t)-2\gamma+1}I_{\{Y\tilde{\boldsymbol{X}}^T(\boldsymbol{\theta}_{0\gamma}+t)\geq\gamma\}} \\
&\quad - (1-Y\tilde{\boldsymbol{X}}^T\boldsymbol{\theta}_{0\gamma})I_{\{Y\widetilde{\boldsymbol{X}}^T\boldsymbol{\theta}_{0\gamma}<\gamma\}} - \frac{(1-\gamma)^2}{Y\tilde{\boldsymbol{X}}^T\boldsymbol{\theta}_{0\gamma}-2\gamma+1}I_{\{Y\tilde{\boldsymbol{X}}^T\boldsymbol{\theta}_{0\gamma}\geq\gamma\}} \\
&= M(\boldsymbol{\theta}_{0\gamma})^T t + R(Z,t),
\end{aligned}$$

where

$$\begin{aligned}
M(\boldsymbol{\theta}_{0\gamma}) &= -Y\tilde{\boldsymbol{X}}^T I_{\{Yf(\tilde{\boldsymbol{X}}^T;\boldsymbol{\theta}_{0\gamma})<\gamma\}} - \frac{(1-\gamma)^2 Y\tilde{\boldsymbol{X}}^T}{(Yf(\tilde{\boldsymbol{X}}^T;\boldsymbol{\theta}_{0\gamma})-2\gamma+1)^2}I_{\{Yf(\tilde{\boldsymbol{X}}^T;\boldsymbol{\theta}_{0\gamma})\geq\gamma\}}; \\
R(Z,t) &= \Big(1-Yf(\boldsymbol{X};\boldsymbol{\theta}_{0\gamma}+t)\Big)\Big[I_{\{Yf(\tilde{\boldsymbol{X}}^T;\boldsymbol{\theta}_{0\gamma}+t)<\gamma\}} - I_{\{Yf(\tilde{\boldsymbol{X}}^T;\boldsymbol{\theta}_{0\gamma})<\gamma\}}\Big] + \frac{(1-\gamma)^2 I_{\{Yf(\tilde{\boldsymbol{X}}^T;\boldsymbol{\theta}_{0\gamma}+t)\geq\gamma\}}}{Yf(\tilde{\boldsymbol{X}}^T;\boldsymbol{\theta}_{0\gamma}+t)-2\gamma+1} \\
&\quad - \Bigg[\frac{(1-\gamma)^2}{Yf(\tilde{\boldsymbol{X}}^T;\boldsymbol{\theta}_{0\gamma})-2\gamma+1} - \frac{(1-\gamma)^2 Yf(\boldsymbol{X},t)}{Yf(\tilde{\boldsymbol{X}}^T;\boldsymbol{\theta}_{0\gamma})-2\gamma+1}\Bigg]I_{\{Yf(\tilde{\boldsymbol{X}}^T;\boldsymbol{\theta}_{0\gamma})\geq\gamma\}}.
\end{aligned}$$

It is easy to check that $E(M(\boldsymbol{\theta}_{0\gamma})) = \nabla_{\boldsymbol{\theta}}\mathcal{R}_\gamma(\boldsymbol{\theta})|_{\boldsymbol{\theta}=\boldsymbol{\theta}_{0\gamma}}$,

$$E[R(Z,t)] = \frac{1}{2}t^T H(\boldsymbol{\theta}_{0\gamma})t + o(\|t\|^2) \ \text{ and } \ E[R^2(Z,t)] = O(\|t\|^3).$$

The remaining arguments follow exactly from the proof of Theorem 1. ■

## S.7. Proof of Lemma 1

In the proof of Corollary 2, we showed that for any $\gamma \in [0,1]$,

$$\sqrt{n}(\widehat{\boldsymbol{\theta}}_\gamma - \boldsymbol{\theta}_{0\gamma}) = -n^{-1/2} H(\boldsymbol{\theta}_{0\gamma})^{-1} \sum_{i=1}^n M_i(\boldsymbol{\theta}_{0\gamma}) + o_P(1); \tag{S.12}$$

$$\sqrt{n}(\widehat{\mathcal{D}}_\gamma - D_{0\gamma}) = n^{-1/2} \sum_{i=1}^n \psi_{i\gamma} + o_P(1), \tag{S.13}$$

where $\psi_{i\gamma} = \frac{1}{2}|Y_i - \text{sign}\{f(\boldsymbol{X}_i;\boldsymbol{\theta}_{0\gamma})\}| - D_{0\gamma} - \dot{d}(\boldsymbol{\theta}_{0\gamma})^T H(\boldsymbol{\theta}_{0\gamma})^{-1} M_i(\boldsymbol{\theta}_{0\gamma})$. In addition, ($S.12$) and ($S.13$) converge to normal distributions.

Next, we show that the right hand sides of ($S.12$) and ($S.13$) are uniformly bounded over $\gamma \in [0,1]$. Denoting the $L_1$ norm as $\|\cdot\|_1$, we have

$$\begin{aligned}
&\sup_{\gamma\in[0,1]} \Big\| M_i(\boldsymbol{\theta}_{0\gamma}) \Big\|_1 \\
\leq\ & \sup_{\gamma\in[0,1]} \Big\| -Y_i \tilde{\boldsymbol{X}}_i I_{(Y_i f(\boldsymbol{X}_i;\boldsymbol{\theta}_{0\gamma})<\gamma)} \Big\|_1 + \sup_{\gamma\in[0,1]} \left\| \frac{(1-\gamma)^2 Y_i \tilde{\boldsymbol{X}}_i I_{(Y_i f(\boldsymbol{X}_i;\boldsymbol{\theta}_{0\gamma})\geq\gamma)}}{\Big(Y_i f(\boldsymbol{X}_i;\boldsymbol{\theta}_{0\gamma}) - 2\gamma + 1\Big)^2} \right\|_1 \\
\leq\ & 2 \Big\| \tilde{\boldsymbol{X}}_i \Big\|_1 .
\end{aligned} \tag{S.14}$$

In addition, $\lambda_{\max}(H(\boldsymbol{\theta}_{0\gamma})) \leq c_2$ in Assumption (B1) implies that each component of the Hessian matrix is uniformly bounded since $\|H(\boldsymbol{\theta}_{0\gamma})\|_{\max} \leq \|H(\boldsymbol{\theta}_{0\gamma})\|_2 = \lambda_{\max}(H(\boldsymbol{\theta}_{0\gamma}))$. This combining with ($S.14$) and Central Limit Theorem leads to

$$\sup_{\gamma\in[0,1]} \Big\| \sqrt{n}(\widehat{\boldsymbol{\theta}}_\gamma - \boldsymbol{\theta}_{0\gamma}) \Big\|_1 = O_P(1). \tag{S.15}$$

Similarly,

$$\begin{aligned}
&\sup_{\gamma\in[0,1]} \Big| \psi_{i\gamma} \Big| \\
\leq\ & \sup_{\gamma\in[0,1]} \frac{1}{2} |Y_i - \text{sign}(\tilde{\boldsymbol{X}}_i^T \boldsymbol{\theta}_{0\gamma})| + \sup_{\gamma\in[0,1]} |D_{0\gamma}| + \sup_{\gamma\in[0,1]} \Big| \dot{d}(\boldsymbol{\theta}_{0\gamma})^T H(\boldsymbol{\theta}_{0\gamma})^{-1} M_i(\boldsymbol{\theta}_{0\gamma}) \Big| \\
\leq\ & 1 + 1 + \sup_{\gamma\in[0,1]} \Big\| \dot{d}(\boldsymbol{\theta}_{0\gamma}) \Big\|_1 \sup_{\gamma\in[0,1]} \Big\| H(\boldsymbol{\theta}_{0\gamma})^{-1} \Big\|_{\max} \sup_{\gamma\in[0,1]} \Big\| M_i(\boldsymbol{\theta}_{0\gamma}) \Big\|_1 \\
\leq\ & 2 + c_3 \|\tilde{\boldsymbol{X}}_i\|_1,
\end{aligned} \tag{S.16}$$

where $c_3$ in ($S.16$) is a constant according to $\|H(\boldsymbol{\theta}_{0\gamma})^{-1}\|_{\max} \leq \|H(\boldsymbol{\theta}_{0\gamma})^{-1}\|_2 = 1/\lambda_{\min}(H(\boldsymbol{\theta}_{0\gamma})) \leq 1/c_1$ from Assumption (B1), and

$$\|\dot{d}(\boldsymbol{\theta}_{0\gamma})\|_1 \leq 4 \Big\| \nabla E\Big( I_{(Y_i f(\boldsymbol{X}_i;\boldsymbol{\theta}_{0\gamma})<0)} \Big) \Big\|_1 \leq 4\delta(-Y_i \boldsymbol{\theta}_{0\gamma}^T \tilde{\boldsymbol{X}}_i) \|\tilde{\boldsymbol{X}}_i\|_1 = 0 \ \text{ a.s.}$$

with $\delta(z)=0$ for $z\neq 0$ and $\infty$ at $z=0$. So ($S.16$) leads to

$$\sup_{\gamma\in[0,1]}\sqrt{n}\left|\widehat{\mathcal{D}}_{\gamma}-D_{0\gamma}\right|=O_P(1). \tag{S.17}$$

In the end, the definitions of $\gamma_0^*$ and $\widehat{\gamma}_0^*$ imply that

$$D_{0\gamma_0^*}-D_{0\widehat{\gamma}_0^*}\leq 0 \quad and \quad \widehat{\mathcal{D}}_{\widehat{\gamma}_0^*}-\widehat{\mathcal{D}}_{\gamma_0^*}\leq 0. \tag{S.18}$$

Therefore, we have $D_{0\gamma_0^*}-\widehat{\mathcal{D}}_{\widehat{\gamma}_0^*}=D_{0\gamma_0^*}-D_{0\widehat{\gamma}_0^*}+D_{0\widehat{\gamma}_0^*}-\widehat{\mathcal{D}}_{\widehat{\gamma}_0^*}\leq D_{0\widehat{\gamma}_0^*}-\widehat{\mathcal{D}}_{\widehat{\gamma}_0^*}=O_P(n^{-1/2})$ based on ($S.17$) and ($S.18$). Using similar arguments, we have $\widehat{\mathcal{D}}_{\widehat{\gamma}_0^*}-D_{0\gamma_0^*}\leq O_P(n^{-1/2})$. The above discussions imply that $\left|\widehat{\mathcal{D}}_{\widehat{\gamma}_0^*}-D_{0\gamma_0^*}\right|=O_P(n^{-1/2})$. This concludes the proof of Lemma 1. ■

## S.8. Lemma 3

The following Lemma will be used in the proof of Lemma 4.

**Lemma 3** *The generalization error* $D_{0\gamma}=\frac{1}{2}E|Y_0-sign\{\tilde{\boldsymbol{X}}_0^T\widehat{\boldsymbol{\theta}}_\gamma\}|$ *is continuous w.r.t.* $\gamma$ *a.s.*

**Proof of Lemma 3:** The discontinuity of sign function happens only at $\tilde{\boldsymbol{X}}_0^T\widehat{\boldsymbol{\theta}}_\gamma=0$, which is assumed to have probability zero. Hence, it is sufficient to show $\widehat{\boldsymbol{\theta}}_\gamma$ is continuous in $\gamma$ by dominated convergence theorem. Recall that $\widehat{\boldsymbol{\theta}}_\gamma=\arg\min_{\boldsymbol{\theta}\in R^{d+1}}O_{n\gamma}(\boldsymbol{\theta})$ with

$$O_{n\gamma}(\boldsymbol{\theta})=\frac{1}{n}\sum_{i=1}^{n}L_\gamma\Big(y_i(\boldsymbol{w}^T\boldsymbol{x}_i+b)\Big)+\frac{\lambda_n\boldsymbol{w}^T\boldsymbol{w}}{2}.$$

Note that $O_{n\gamma}(\boldsymbol{\theta})$ is continuous w.r.t. $\gamma$ due to the continuity of $L_\gamma(u)$ w.r.t. $\gamma$. Then, for any sequence $\gamma_n\to\gamma_{00}$ with $\gamma_{00}\in[0,1]$, continuous mapping theorem implies that $|O_{n\gamma_n}(\boldsymbol{\theta})-O_{n\gamma_{00}}(\boldsymbol{\theta})|<\delta$ for any $\delta>0$ when $n$ is sufficiently large. Denote $\widehat{\boldsymbol{\theta}}_{\gamma_{00}}=\arg\min_{\boldsymbol{\theta}}O_{n\gamma_{00}}(\boldsymbol{\theta})$ and $\mathcal{G}=\{\boldsymbol{\theta}:\|\boldsymbol{\theta}-\widehat{\boldsymbol{\theta}}_{\gamma_{00}}\|\leq\epsilon\}$. For each fixed $\epsilon$, we construct

$$\delta=\frac{\min_{\boldsymbol{\theta}\in R^{d+1}\backslash\mathcal{G}}O_{n\gamma_{00}}(\boldsymbol{\theta})-O_{n\gamma_{00}}(\widehat{\boldsymbol{\theta}}_{\gamma_{00}})}{2}.$$

Then we have

$$\begin{aligned}O_{n\gamma_{00}}(\widehat{\boldsymbol{\theta}}_{\gamma_{00}}) &= \min_{\boldsymbol{\theta}\in R^{d+1}\backslash\mathcal{G}}O_{n\gamma_{00}}(\boldsymbol{\theta})-2\delta\\ &< \min_{\boldsymbol{\theta}\in R^{d+1}\backslash\mathcal{G}}O_{n\gamma_{00}}(\boldsymbol{\theta})+O_{n\gamma_n}(\boldsymbol{\theta})-O_{n\gamma_{00}}(\boldsymbol{\theta})-\delta\\ &\leq O_{n\gamma_n}(\boldsymbol{\theta})-\delta,\end{aligned}$$

which is true for any $\boldsymbol{\theta}\in R^{d+1}$. Therefore,

$$O_{n\gamma_{00}}(\widehat{\boldsymbol{\theta}}_{\gamma_{00}})<\min_{\boldsymbol{\theta}\in R^{d+1}\backslash\mathcal{G}}O_{n\gamma_n}(\boldsymbol{\theta})-\delta. \tag{S.19}$$

On the other hand, $|O_{n\gamma_n}(\boldsymbol{\theta})-O_{n\gamma_{00}}(\boldsymbol{\theta})|<\delta$ implies that $O_{n\gamma_n}(\widehat{\boldsymbol{\theta}}_{\gamma_{00}})-O_{n\gamma_{00}}(\widehat{\boldsymbol{\theta}}_{\gamma_{00}})<\delta$ and hence $\min_{\boldsymbol{\theta}\in R^{d+1}} O_{n\gamma_n}(\boldsymbol{\theta}) < O_{n\gamma_{00}}(\widehat{\boldsymbol{\theta}}_{\gamma_{00}}) + \delta$. This combining with (S.19) leads to

$$\min_{\boldsymbol{\theta}\in R^{d+1}} O_{n\gamma_n}(\boldsymbol{\theta}) < \min_{\boldsymbol{\theta}\in R^{d+1}\setminus\mathcal{G}} O_{n\gamma_n}(\boldsymbol{\theta}).$$

Therefore, $\arg\min_{\boldsymbol{\theta}\in R^{d+1}} O_{n\gamma_n}(\boldsymbol{\theta}) \in \mathcal{G}$, and hence $\widehat{\boldsymbol{\theta}}_\gamma$ is continuous at $\gamma_{00}$. Note that $\epsilon$ can be made arbitrarily small and $\gamma_{00}$ is an arbitrary element within $[0,1]$. This concludes Lemma 3. ■

## S.9. Lemma 4

Lemma 4 shows the (element-wise) asymptotic equivalence between $\Lambda_0$ and $\widehat{\Lambda}_0$. It will be used in the proof of Theorem 3.

**Lemma 4** *Suppose that the assumptions in Lemma 1 hold. We have, as* $n \to \infty$, *(i) for any* $\widehat{\gamma} \in \widehat{\Lambda}_0$, *there exists a* $\gamma \in \Lambda_0$ *such that* $\widehat{\gamma} \xrightarrow{P} \gamma$; *(ii) for any* $\gamma \in \Lambda_0$, *there exists a* $\widehat{\gamma} \in \widehat{\Lambda}_0$ *satisfying* $\widehat{\gamma} \xrightarrow{P} \gamma$.

**Proof of Lemma 4:** Our proof consists of two steps. In the first step, for any $\widehat{\gamma} \in \widehat{\Lambda}_0$ with $\widehat{\gamma} \xrightarrow{P} \gamma$, we have

$$\begin{aligned} D_{0\gamma} - D_{0\gamma_0^*} &= (D_{0\gamma} - D_{0\widehat{\gamma}}) + (D_{0\widehat{\gamma}} - \widehat{\mathcal{D}}_{\widehat{\gamma}}) + (\widehat{\mathcal{D}}_{\widehat{\gamma}} - \widehat{\mathcal{D}}_{\gamma_0^*}) + (\widehat{\mathcal{D}}_{\gamma_0^*} - D_{0\gamma_0^*}) \\ &= I + II + III + IV. \end{aligned}$$

Obviously, we have $I = o_P(1)$ according to continuous mapping theorem and Lemma 3, and $II, IV = o_P(1)$ due to (S.17). As for $III$, we have $III \le \widehat{\mathcal{D}}_{\widehat{\gamma}_0^*} - \widehat{\mathcal{D}}_{\gamma_0^*} + n^{-1/2}\phi_{\widehat{\gamma},\widehat{\gamma}_0^*;\alpha/2} \le o_P(1)$ since $\widehat{\gamma} \in \widehat{\Lambda}_0$ defined in (15). The above discussions lead to the conclusion that $D_{0\gamma} - D_{0\gamma_0^*} \le o_P(1)$. Therefore, we have $P(\gamma \in \Lambda_0) \ge P(D_{0\gamma} - D_{0\gamma_0^*} \le 0) \to 1$.

In the second step, we apply the contradiction argument. Assume there exists some $\gamma \in \Lambda_0$ such that $\widehat{\gamma} \notin \widehat{\Lambda}_0$ for any $\widehat{\gamma} \xrightarrow{P} \gamma$. The above assumption directly implies that $\widehat{\mathcal{D}}_{\widehat{\gamma}} - \widehat{\mathcal{D}}_{\widehat{\gamma}_0^*} > o_P(1)$. The analysis in the first step further implies that there exists some $\gamma^* \in \Lambda_0$, i.e., $D_{0\gamma^*} = D_{0\gamma_0^*}$, with probability tending to one such that $\widehat{\gamma}_0^* \xrightarrow{P} \gamma^*$. Then, we have

$$\begin{aligned} D_{0\gamma} - D_{0\gamma^*} &= (D_{0\gamma} - D_{0\widehat{\gamma}}) + (D_{0\widehat{\gamma}} - \widehat{\mathcal{D}}_{\widehat{\gamma}}) + (\widehat{\mathcal{D}}_{\widehat{\gamma}} - \widehat{\mathcal{D}}_{\widehat{\gamma}_0^*}) + (\widehat{\mathcal{D}}_{\widehat{\gamma}_0^*} - D_{0\gamma^*}) \\ &= I + II + III' + IV'. \end{aligned}$$

Recall that $I, II = o_P(1)$ and $III' > o_P(1)$ as shown in the above. We also have $IV' = o_P(1)$ due to (S.17) and the fact that $\widehat{\gamma}_0^* \xrightarrow{P} \gamma^*$. In summary, we have $D_{0\gamma} - D_{0\gamma^*} > o_P(1)$, which contradicts the definition of $\gamma$. This concludes the proof of Lemma 4. ■

## S.10. Proof of Theorem 3

The proof consists of two major steps. In the first step, we show that

$$\sup_{\gamma\in[0,1]} n\left|\widehat{DBI}(S(\boldsymbol{X};\widehat{\boldsymbol{\theta}}_\gamma)) - DBI(S(\boldsymbol{X};\widehat{\boldsymbol{\theta}}_\gamma))\right| \to 0. \tag{S.20}$$

Denote $\overline{DBI}(S(\boldsymbol{X};\widehat{\boldsymbol{\theta}}_\gamma)) = \frac{1}{n}\sum_{i=1}^n \widetilde{\boldsymbol{x}}_{i}{}_{(-d)}^{\dagger T} Var(\widehat{\boldsymbol{\eta}}_\gamma^\dagger)\widetilde{\boldsymbol{x}}_{i}{}_{(-d)}^{\dagger}$, where $\widetilde{\boldsymbol{x}}_{i}{}_{(-d)}^{\dagger} = (1, (R_\gamma \boldsymbol{x}_i)_{(-d)}^T)^T$ and $R_\gamma$ is the transformation matrix associated with the loss function $L_\gamma$. Then we have

$$\begin{aligned}
&\sup_{\gamma\in[0,1]} n\Big|\widehat{DBI}(S(\boldsymbol{X};\widehat{\boldsymbol{\theta}}_\gamma)) - DBI(S(\boldsymbol{X};\widehat{\boldsymbol{\theta}}_\gamma))\Big| \\
\leq\ & \sup_{\gamma\in[0,1]} n\Big|\widehat{DBI}(S(\boldsymbol{X};\widehat{\boldsymbol{\theta}}_\gamma)) - \overline{DBI}(S(\boldsymbol{X};\widehat{\boldsymbol{\theta}}_\gamma))\Big| + \sup_{\gamma\in[0,1]} n\Big|\overline{DBI}(S(\boldsymbol{X};\widehat{\boldsymbol{\theta}}_\gamma)) - DBI(S(\boldsymbol{X};\widehat{\boldsymbol{\theta}}_\gamma))\Big|.
\end{aligned} \tag{S.21}$$

Next we show each summand in (S.21) converges to 0. For the first summand, we have

$$\begin{aligned}
&\sup_{\gamma\in[0,1]} n\Big|\widehat{DBI}(S(\boldsymbol{X};\widehat{\boldsymbol{\theta}}_\gamma)) - \overline{DBI}(S(\boldsymbol{X};\widehat{\boldsymbol{\theta}}_\gamma))\Big| \\
=\ & \sup_{\gamma\in[0,1]} n\Big|\frac{1}{n}\sum_{i=1}^n \widetilde{\boldsymbol{x}}_{i}{}_{(-d)}^{\dagger T}\widehat{Var}(\widehat{\boldsymbol{\eta}}_\gamma^\dagger)\widetilde{\boldsymbol{x}}_{i}{}_{(-d)}^{\dagger} - \frac{1}{n}\sum_{i=1}^n \widetilde{\boldsymbol{x}}_{i}{}_{(-d)}^{\dagger T} Var(\widehat{\boldsymbol{\eta}}_\gamma^\dagger)\widetilde{\boldsymbol{x}}_{i}{}_{(-d)}^{\dagger}\Big| \\
=\ & \sup_{\gamma\in[0,1]} \Big|\sum_{i=1}^n \widetilde{\boldsymbol{x}}_{i}{}_{(-d)}^{\dagger T}[(\widehat{Var}(\widehat{\boldsymbol{\eta}}_\gamma^\dagger) - Var(\widehat{\boldsymbol{\eta}}_\gamma^\dagger))]\widetilde{\boldsymbol{x}}_{i}{}_{(-d)}^{\dagger}\Big|,
\end{aligned} \tag{S.22}$$

where

$$Var(\widehat{\boldsymbol{\eta}}_\gamma^\dagger) = \frac{\Sigma_{0\gamma,(-d)}^\dagger}{n(w_{\gamma,d}^\dagger)^2} \quad\text{and}\quad \widehat{Var}(\widehat{\boldsymbol{\eta}}_\gamma^\dagger) = \frac{\widehat{\Sigma}_{\gamma,(-d)}^\dagger}{n(\widehat{w}_{\gamma,d}^\dagger)^2}.$$

Here, $\widehat{w}_{\gamma,d}^\dagger$ is the last dimension of $\widehat{\boldsymbol{\theta}}_\gamma^*$. Since $\widehat{w}_{\gamma,d}^\dagger$ follows the normal distribution with mean $w_{\gamma,d}^\dagger$ and variance converging to 0, we have $\widehat{w}_{\gamma,d}^\dagger = w_{\gamma,d}^\dagger + o_P(1)$, and hence $(\widehat{w}_{\gamma,d}^\dagger)^2 = (w_{\gamma,d}^\dagger)^2 + o_P(1)$ due to the boundedness of $w_{\gamma,d}^\dagger$. In addition, uniform law of large numbers implies that each component of $\widehat{\Sigma}_\gamma^\dagger - \Sigma_{0\gamma}^\dagger$ uniformly converges to 0 w.r.t. $\gamma$, because each element of $\widehat{\Sigma}_\gamma^\dagger$ is continuous w.r.t. $\gamma$ (by similar arguments as in Lemma 3). Therefore, we have

$$\begin{aligned}
n\Big[\widehat{Var}(\widehat{\boldsymbol{\eta}}_\gamma^\dagger) - Var(\widehat{\boldsymbol{\eta}}_\gamma^\dagger)\Big] &= \frac{\widehat{\Sigma}_{\gamma,(-d)}^\dagger}{(\widehat{w}_{\gamma,d}^\dagger)^2} - \frac{\Sigma_{0\gamma,(-d)}^\dagger}{(w_{\gamma,d}^\dagger)^2} \\
&= \frac{\widehat{\Sigma}_{\gamma,(-d)}^\dagger - \Sigma_{0\gamma,(-d)}^\dagger}{(w_{\gamma,d}^\dagger)^2 + o_P(1)} - \frac{\Sigma_{0\gamma,(-d)}^\dagger o_P(1)}{(w_{\gamma,d}^\dagger)^2[(w_{\gamma,d}^\dagger)^2 + o_P(1)]},
\end{aligned} \tag{S.23}$$

where the second term in ($S.23$) uniformly converges to 0 due to Assumption (B1) and the boundedness of $w_{\gamma,d}^\dagger$. Therefore, each element of ($S.23$) uniformly converges to 0, which implies that ($S.22$) converges to 0.

As for the second summand of (S.21), we again apply uniform law of large numbers to show

$$\sup_{\gamma\in[0,1]} n\Big|\overline{DBI}(S(\boldsymbol{X};\widehat{\boldsymbol{\theta}}_\gamma)) - DBI(S(\boldsymbol{X};\widehat{\boldsymbol{\theta}}_\gamma))\Big| \to 0.$$

Note that $\tilde{\boldsymbol{X}}^{\dagger T}_{(-d)} Var(\widehat{\boldsymbol{\eta}}_\gamma)^\dagger \tilde{\boldsymbol{X}}^{\dagger}_{(-d)}$ is continuous w.r.t. $\gamma$ by similar arguments as in Lemma 3, and

$$n\Big|\tilde{\boldsymbol{X}}^{\dagger T}_{(-d)} Var(\widehat{\boldsymbol{\eta}}^\dagger_\gamma)\tilde{\boldsymbol{X}}^{\dagger}_{(-d)}\Big| = \Big|(1,(R_\gamma\boldsymbol{x})^T_{(-d)})^T nVar(\widehat{\boldsymbol{\eta}}^\dagger_\gamma)(1,(R_\gamma\boldsymbol{x})^T_{(-d)})\Big| \le c_4\Big|1+\boldsymbol{x}^T_{(-d)}\boldsymbol{x}_{(-d)}\Big| \le c_5,$$

where the first inequality holds because each component of $nVar(\widehat{\boldsymbol{\eta}}^\dagger_\gamma)$ is uniformly bounded due to the boundedness of $w^\dagger_{\gamma,d}$ and Assumption (B1). Then the uniform law of large number implies

$$\begin{aligned}
&\sup_{\gamma\in[0,1]} n\Big|\overline{DBI}(S(\boldsymbol{X};\widehat{\boldsymbol{\theta}}_\gamma)) - DBI(S(\boldsymbol{X};\widehat{\boldsymbol{\theta}}_\gamma))\Big| \\
= &\sup_{\gamma\in[0,1]} \Big|\frac{1}{n}\sum_{i=1}^n \tilde{\boldsymbol{x}}_{i}{}^{\dagger T}_{(-d)}(w^\dagger_{\gamma,d})^{-2}\Sigma^\dagger_{0\gamma,(-d)}(\widehat{\boldsymbol{\eta}}^\dagger_\gamma)\tilde{\boldsymbol{x}}_{i}{}^{\dagger}_{(-d)} - E\Big(\tilde{\boldsymbol{X}}^{\dagger T}_{(-d)}(w^\dagger_{\gamma,d})^{-2}\Sigma^\dagger_{0\gamma,(-d)}\tilde{\boldsymbol{X}}^{\dagger}_{(-d)}\Big)\Big| \\
\to &\ 0.
\end{aligned} \tag{S.24}$$

Combining (*S*.22) and (*S*.24) leads to (*S*.20).

In the second step of the proof, we show $n(\widehat{DBI}(S(\boldsymbol{X};\widehat{\boldsymbol{\theta}}_{\widehat{\gamma}_0})) - DBI(S(\boldsymbol{X};\widehat{\boldsymbol{\theta}}_{\gamma_0}))) \le o_P(1)$ and $n(DBI(S(\boldsymbol{X};\widehat{\boldsymbol{\theta}}_{\gamma_0})) - \widehat{DBI}(S(\boldsymbol{X};\widehat{\boldsymbol{\theta}}_{\widehat{\gamma}_0}))) \le o_P(1)$, from which the desirable result (20) follows.

Firstly, we prove

$$n\Big(\widehat{DBI}(S(\boldsymbol{X};\widehat{\boldsymbol{\theta}}_{\widehat{\gamma}_0})) - DBI(S(\boldsymbol{X};\widehat{\boldsymbol{\theta}}_{\gamma_0}))\Big) \le o_P(1).$$

Denote $\widehat{\gamma}_0^\sharp = \arg\min_{\gamma\in\widehat{\Lambda}_0} DBI(S(\boldsymbol{X};\widehat{\boldsymbol{\theta}}_\gamma))$. For $\gamma_0$ defined in (19), Theorem 4 implies that there exists a $\widehat{\gamma}_0^\triangle \in \widehat{\Lambda}_0$ such that $\widehat{\gamma}_0^\triangle \xrightarrow{P} \gamma_0$, then we have

$$\begin{aligned}
&n\Big(\widehat{DBI}(S(\boldsymbol{X};\widehat{\boldsymbol{\theta}}_{\widehat{\gamma}_0})) - DBI(S(\boldsymbol{X};\widehat{\boldsymbol{\theta}}_{\gamma_0}))\Big) \\
= &\ n\Big(\widehat{DBI}(S(\boldsymbol{X};\widehat{\boldsymbol{\theta}}_{\widehat{\gamma}_0})) - DBI(S(\boldsymbol{X};\widehat{\boldsymbol{\theta}}_{\widehat{\gamma}_0^\sharp}))\Big) + n\Big(DBI(S(\boldsymbol{X};\widehat{\boldsymbol{\theta}}_{\widehat{\gamma}_0^\sharp})) - DBI(S(\boldsymbol{X};\widehat{\boldsymbol{\theta}}_{\widehat{\gamma}_0^\triangle}))\Big) \\
&\quad + n\Big(DBI(S(\boldsymbol{X};\widehat{\boldsymbol{\theta}}_{\widehat{\gamma}_0^\triangle})) - DBI(S(\boldsymbol{X};\widehat{\boldsymbol{\theta}}_{\gamma_0}))\Big) \\
\le &\ n\Big(\widehat{DBI}(S(\boldsymbol{X};\widehat{\boldsymbol{\theta}}_{\widehat{\gamma}_0^\sharp})) - DBI(S(\boldsymbol{X};\widehat{\boldsymbol{\theta}}_{\widehat{\gamma}_0^\sharp}))\Big) + n\Big(DBI(S(\boldsymbol{X};\widehat{\boldsymbol{\theta}}_{\widehat{\gamma}_0^\sharp})) - DBI(S(\boldsymbol{X};\widehat{\boldsymbol{\theta}}_{\widehat{\gamma}_0^\triangle}))\Big) \\
&\quad + n\Big(DBI(S(\boldsymbol{X};\widehat{\boldsymbol{\theta}}_{\widehat{\gamma}_0^\triangle})) - DBI(S(\boldsymbol{X};\widehat{\boldsymbol{\theta}}_{\gamma_0}))\Big) \\
\le &\ \sup_{\gamma\in\widehat{\Lambda}_0} n\Big|\widehat{DBI}(S(\boldsymbol{X};\widehat{\boldsymbol{\theta}}_\gamma)) - DBI(S(\boldsymbol{X};\widehat{\boldsymbol{\theta}}_\gamma))\Big| + o_P(1) \\
\le &\ o_P(1),
\end{aligned} \tag{S.25}$$

where $\widehat{DBI}(S(\boldsymbol{X};\widehat{\boldsymbol{\theta}}_{\widehat{\gamma}_0})) \le \widehat{DBI}(S(\boldsymbol{X};\widehat{\boldsymbol{\theta}}_{\widehat{\gamma}_0^\sharp}))$ according to (18), $DBI(S(\boldsymbol{x};\widehat{\boldsymbol{\theta}}_{\widehat{\gamma}_0^\sharp})) \le DBI(S(\boldsymbol{x};\widehat{\boldsymbol{\theta}}_{\widehat{\gamma}_0^\triangle}))$ due to $\widehat{\gamma}_0^\sharp \in \widehat{\Lambda}_0$, $DBI(S(\boldsymbol{X};\widehat{\boldsymbol{\theta}}_{\widehat{\gamma}_0^\triangle})) - DBI(S(\boldsymbol{X};\widehat{\boldsymbol{\theta}}_{\gamma_0})) = o_P(n^{-1})$ according to $\widehat{\gamma}_0^\triangle \xrightarrow{P} \gamma_0$ and continuous mapping theorem. All these together with (*S*.20) lead to (*S*.25).

Secondly, we prove

$$n(DBI(S(\boldsymbol{X};\widehat{\boldsymbol{\theta}}_{\gamma_0})) - \widehat{DBI}(S(\boldsymbol{X};\widehat{\boldsymbol{\theta}}_{\widehat{\gamma}_0}))) \le o_P(1).$$

Denote $\widetilde{\gamma}_0 = \arg\min_{\gamma\in\Lambda_0} \widehat{DBI}(S(\boldsymbol{X};\widehat{\boldsymbol{\theta}}_\gamma))$. For $\widehat{\gamma}_0$ defined in (18), Lemma 4 implies that there exists $\widetilde{\gamma}_0^\sharp \in \Lambda_0$ such that $\widehat{\gamma}_0 \xrightarrow{P} \widetilde{\gamma}_0^\sharp$, then we have

$$
\begin{aligned}
& n\Big(DBI(S(\boldsymbol{X};\widehat{\boldsymbol{\theta}}_{\gamma_0})) - \widehat{DBI}(S(\boldsymbol{X};\widehat{\boldsymbol{\theta}}_{\widehat{\gamma}_0}))\Big) \\
\leq\ & n\Big(DBI(S(\boldsymbol{X};\widehat{\boldsymbol{\theta}}_{\widehat{\gamma}_0})) - \widehat{DBI}(S(\boldsymbol{X};\widehat{\boldsymbol{\theta}}_{\widehat{\gamma}_0}))\Big) + n\Big(\widehat{DBI}(S(\boldsymbol{X};\widehat{\boldsymbol{\theta}}_{\widetilde{\gamma}_0})) - \widehat{DBI}(S(\boldsymbol{X};\widehat{\boldsymbol{\theta}}_{\widetilde{\gamma}_0^\sharp}))\Big) \\
& \quad + n\Big(\widehat{DBI}(S(\boldsymbol{X};\widehat{\boldsymbol{\theta}}_{\widetilde{\gamma}_0^\sharp})) - \widehat{DBI}(S(\boldsymbol{X};\widehat{\boldsymbol{\theta}}_{\widehat{\gamma}_0}))\Big) \\
\leq\ & \sup_{\gamma\in\Lambda_0} n\Big|\widehat{DBI}(S(\boldsymbol{X};\widehat{\boldsymbol{\theta}}_{\gamma})) - DBI(S(\boldsymbol{X};\widehat{\boldsymbol{\theta}}_{\gamma}))\Big| + o_P(1) \leq o_P(1),
\end{aligned}
\tag{S.26}
$$

where $DBI(S(\boldsymbol{X};\widehat{\boldsymbol{\theta}}_{\gamma_0})) \leq DBI(S(\boldsymbol{X};\widehat{\boldsymbol{\theta}}_{\widehat{\gamma}_0}))$ by the definition of $\gamma_0$, $\widehat{DBI}(S(\boldsymbol{X};\widehat{\boldsymbol{\theta}}_{\widetilde{\gamma}_0})) \leq \widehat{DBI}(S(\boldsymbol{X};\widehat{\boldsymbol{\theta}}_{\widetilde{\gamma}_0^\sharp}))$ due to the definition of $\widetilde{\gamma}_0$, and $\widehat{DBI}(S(\boldsymbol{X};\widehat{\boldsymbol{\theta}}_{\widetilde{\gamma}_0^\sharp})) - \widehat{DBI}(S(\boldsymbol{X};\widehat{\boldsymbol{\theta}}_{\widehat{\gamma}_0})) = o_P(n^{-1})$ according to $\widehat{\gamma}_0 \xrightarrow{P} \widetilde{\gamma}_0^\sharp$ and continuous mapping theorem.

Consequently, combining ($S$.25) and ($S$.26) leads to $n\Big|\widehat{DBI}(S(\boldsymbol{X};\widehat{\boldsymbol{\theta}}_{\widehat{\gamma}_0})) - DBI(S(\boldsymbol{X};\widehat{\boldsymbol{\theta}}_{\gamma_0}))\Big| \to 0$, which concludes the proof of Theorem 3. ■

### S.11. Notation Table S1

Table S1: Important notation, its meaning, and where it first appears.

| Notation | Meaning | Section No. |
|---|---|---|
| $\boldsymbol{x}$ | input variable | 2 |
| $\tilde{\boldsymbol{x}}$ | $\tilde{\boldsymbol{x}} = (1, \boldsymbol{x}^T)^T$ | 2 |
| $p(\boldsymbol{x})$ | conditional probability $P(Y = 1 \mid \boldsymbol{X} = \boldsymbol{x})$ | 2 |
| $b$,$\boldsymbol{w}$, $\boldsymbol{\theta}$ | intercept, coefficient and parameter $\boldsymbol{\theta} = (b, \boldsymbol{w}^T)^T$ | 2 |
| $S(\boldsymbol{x}; \boldsymbol{\theta})$ | the decision boundary induced from $\boldsymbol{\theta}$ | 2 |
| $\mathcal{P}(\boldsymbol{X}, Y)$ | joint distribution of $(\boldsymbol{X}, Y)$ | 2 |
| $\mathcal{R}_L$ | risk of loss function $L$ | 2 |
| $b_{0L}$, $\boldsymbol{w}_{0L}$, $\boldsymbol{\theta}_{0L}$ | true intercept, coefficient, and parameter | 2 |
| $(\boldsymbol{x}_i, y_i)$, $\mathcal{D}_n$ | training data $\mathcal{D}_n = \{(\boldsymbol{x}_i, y_i), i = 1, \ldots, n\}$ | 2 |
| $O_{nL}$ | empirical risk of loss function $L$ | 2 |
| $\widehat{b}_L$, $\widehat{\boldsymbol{w}}_L$, $\widehat{\boldsymbol{\theta}}_L$ | estimated intercept, coefficient, and parameter | 2 |
| $D_{0L}$ | GE from loss function L | 3.1 |
| $\widehat{D}_L$, $\widehat{D}(\widehat{\boldsymbol{\theta}}_L)$ | empirical generalization error from loss function L | 3.1 |
| $\widehat{\mathcal{D}}_L$ | K-CV error from loss function L | 3.1 |
| $G(\boldsymbol{\theta}_{0L})$, $H(\boldsymbol{\theta}_{0L})$ | the gradient matrix and Hessian matrix | 3.1 |
| $\mathcal{W}_L$ | $= n^{1/2}(\widehat{\mathcal{D}}_L - D_{0L})$ | 3.1 |
| $D_0(\boldsymbol{\theta})$ | $= \frac{1}{2}E\lvert y_0 - \text{sign}\{f(\boldsymbol{x}_0; \boldsymbol{\theta})\}\rvert$ | 3.1 |
| $\dot{d}(\boldsymbol{\theta})$ | $= \nabla_{\boldsymbol{\theta}} E(\widehat{D}(\boldsymbol{\theta}))$ | 3.1 |
| $D_{0j}$, $\widehat{D}_j$, $\widehat{\mathcal{D}}_j$ | GE, empirical GE, and K-CV error w.r.t $L_j$ | 3.1 |
| $\Delta_{12}$, $\widehat{\Delta}_{12}$ | $\Delta_{12} = D_{02} - D_{01}$; $\widehat{\Delta}_{12} = \widehat{\mathcal{D}}_2 - \widehat{\mathcal{D}}_1$ | 3.1 |
| $\mathcal{W}_j$ | $= n^{1/2}(\widehat{\mathcal{D}}_j - D_{0j})$ | 3.1 |
| $\mathcal{W}_{\Delta_{12}}$ | $= \mathcal{W}_2 - \mathcal{W}_1$ | 3.1 |
| $G_i$ | the random variable generated from Exp(1) | 3.1 |
| $\widehat{\boldsymbol{\theta}}_j^*$, $W_j^*$, $W_{\Delta_{12}}^*$ | the perturbed version of the corresponding terms | 3.1 |
| $\widehat{\boldsymbol{\theta}}_j^{*(r)}$, $W_j^{*(r)}$, $W_{\Delta_{12}}^{*(r)}$ | the corresponding terms in the $r$th replication | 3.1 |
| $\phi_{1,2;\alpha}$ | the $\alpha$th upper percentile of the sequence $W_{\Delta_{12}}^{*(r)}$ | 3.1 |
| $\mathcal{X}_1, \ldots, \mathcal{X}_d$ | the original axes | 3.2 |
| $R_L$ | the transformation matrix induced from loss $L$ | 3.2 |
| $\widehat{b}_L^\dagger$, $\widehat{\boldsymbol{w}}_L^\dagger$, $\widehat{\boldsymbol{\theta}}_L^\dagger$ | transformed estimates of parameters | 3.2 |
| $\Sigma_{0L}^\dagger$, $\widehat{\Sigma}_L^\dagger$ | the covariance matrix and its transformed estimator | 3.2 |
| $\Sigma_{0L,(-d)}^\dagger$, $\widehat{\Sigma}_{L,(-d)}^\dagger$ | removing last row and last column of $\Sigma_{0L}^\dagger$ and $\widehat{\Sigma}_L^\dagger$ | 3.2 |
| $L_\gamma$ | the LUM loss function indexed by $\gamma$ | 4 |
| $\boldsymbol{\theta}_{0\gamma}$, $\widehat{\boldsymbol{\theta}}_\gamma$ | true and estimated parameter from $L_\gamma$ | 4 |
| $\mathcal{R}_\gamma$ | true risk from $L_\gamma$ | 4 |
| $D_{0\gamma}$, $\widehat{\mathcal{D}}_\gamma$ | GE and CV error from $L_\gamma$ | 4 |
| $\gamma_0^*$, $\widehat{\gamma}_0^*$ | LUM index of minimal GE, minimal K-CV error | 4 |
| $\Lambda_0$, $\widehat{\Lambda}_0$ | true and estimated set of potentially good classifiers | 4 |
| $\gamma_0$, $\widehat{\gamma}_0$ | optimal index and its estimate | 4 |